\documentclass[10pt, twocolumn]{article}

\usepackage[margin=0.75in]{geometry}

\usepackage{microtype}
\usepackage{graphicx}
\usepackage{subfigure}
\usepackage{booktabs} 

\usepackage{natbib}

\usepackage{amsmath}
\usepackage{amssymb}
\usepackage{mathtools}
\usepackage{amsthm}

\usepackage{dsfont}
\usepackage{caption}
\usepackage{enumitem}
\usepackage{multirow}
\usepackage{hhline}
\usepackage{comment}

\usepackage{hyperref}
\usepackage[capitalize,noabbrev]{cleveref}

\theoremstyle{plain}

\theoremstyle{definition}

\theoremstyle{remark}

\usepackage{xcolor}
\newcommand{\update}[1]{\textcolor{blue}{#1}}

\DeclareMathOperator{\R}{\mathbb{R}}
\DeclareMathOperator{\E}{\mathbb{E}}

\DeclareMathOperator*{\argmax}{arg\,max}

\DeclareMathOperator{\T}{\mathcal{T}}
\DeclareMathOperator{\StateSpace}{\mathcal{S}}
\DeclareMathOperator{\ActionSpace}{\mathcal{A}}
\DeclareMathOperator{\ReplayBuffer}{\mathcal{D}}
\newcommand{\SAC}{\texttt{SAC}}
\newcommand{\IQL}{\texttt{IQL}}
\newcommand{\PEX}{\texttt{PEX}}
\newcommand{\FT}{\texttt{FT}}
\newcommand{\OFF}{\texttt{OFF}}
\newcommand{\CalQL}{\texttt{CalQL}}
\newcommand{\EQL}{\texttt{EQL}}
\newcommand{\PROTO}{\texttt{PROTO}}

\begin{document}

\title{An Empirical Study on the Effectiveness of \\ Incorporating Offline RL As Online RL Subroutines}
\author{
    Jianhai Su \quad Jinzhu Luo \quad Qi Zhang \\
    University of South Carolina
}
\date{}
\maketitle

\begin{abstract}
We take the novel perspective of incorporating offline RL algorithms as subroutines of tabula rasa online RL.
This is feasible because an online learning agent can repurpose its historical interactions as offline dataset.
We formalize this idea into a framework that accommodates several variants of offline RL incorporation such as final policy recommendation and online fine-tuning.
We further introduce convenient techniques to improve its effectiveness in enhancing online learning efficiency.
Our extensive and systematic empirical analyses show that 
1) the effectiveness of the proposed framework depends strongly on the nature of the task,
2) our proposed techniques greatly enhance its effectiveness,
and 3) existing online fine-tuning methods are overall ineffective, calling for more research therein.
\end{abstract}

\setlength{\tabcolsep}{1mm}

\section{Introduction}\label{sec:intro}
Research in reinforcement learning (RL) has been primarily focusing on two problem settings.
In the {\em online} setting, the agent learns from scratch through a process of repeated interactions with its environment; in the {\em offline} setting, the agent is given a fixed dataset of interactions logging, from which it aims to extract a decision-making policy.
While both settings aim to find good policies, their respective challenges are contrastive:
key to sample-efficient online RL is efficient exploration of its state-action space, while the challenge of offline RL lies in how to effectively estimate and incorporate values for actions that are not in the dataset.
Therefore, research in both settings has developed separate solutions addressing their respective challenges. 

While offline RL is a problem of its own right, its algorithms hold the promise of benefiting the online setting because
1) it is feasible to incorporate offline RL as online subroutines, as the online agent can pause its interaction and repurpose its past interactions as the offline dataset, 
and 2) offline RL algorithms hold the promise of extracting policies better than the one(s) used to collect the dataset and thus the subroutines can significantly enhance learning efficiency. 
Motivated by this key observation, we focus on online RL (and not offline RL) as the problem setting and ask the following question:
{\em How much can online RL benefit from incorporating offline RL algorithms as its subroutines?}  

Although the question seems natural to ask, it has been surprisingly under-studied.
Prior work on online RL has mostly focused on developing ``purely'' online algorithms where the agent performs incremental updates to its internal state (e.g., one or multiple gradient steps to its function approximators), without invoking an offline learning procedure.

Recently, there has been works on combining the offline and online settings. \citet{PEX} introduce a policy expansion method to balance conservatism with exploration, smoothing the offline-to-online transition. \citet{zheng2023adaptive} propose an adaptive framework that enhances policy performance by integrating offline data with online interactions. \citet{lee2022offline} improve sample efficiency using a balanced replay mechanism and a pessimistic Q-ensemble to mitigate distributional shifts. \citet{niu2022trust} propose a framework for enhancing online learning by using offline data to guide online policy learning.
The prior works, however, assume {\em a priori} the existence of a offline dataset and/or a reference policy before starting the online learning process, as we review in detail in Appendix \ref{asec:Extended Related Work}.

In contrast, we focus on the standard online setting of tabula rasa RL, where the agent has to learn from scratch without any dataset or policy given.
We present an extensive and systematic empirical study into the motivating question that makes the following contributions:
\begin{description}[topsep = 0pt, itemsep=0pt, leftmargin=*]
\item  [Formalization.]
We formalize a universal framework consisting of a family of online RL processes that incorporate offline RL algorithms as their subroutines, where the agent will pause at some point to prepare a dataset before invoking an offline RL algorithm.
The introduced online processes vary in their complexity of incorporating offline RL. 
The most straightforward one just calls an offline RL algorithm at the end of interaction to output its final policy. 
In more complicated ones, the offline RL is followed by additional online interactions for fine-tuning the policy.

\item [Robust offline RL incorporation.]
When naively incorporating offline RL as online subroutines, we identify several factors that hinder its effectiveness, including 
(i) low-quality datasets caused by unstable online policy learning, 
(ii) overfitting in offline RL,
and (iii) ineffective online fine-tuning.
Note that (i) and (iii) are not relevant in purely offline RL, while existing work in offline RL has provided little to address (ii).  
We introduce several techniques that are simple yet effective to overcome these issues. 

\item [Empirical analyses.]
To understand when incorporating offline RL is beneficial,
we have performed extensive and systematic empirical analyses to evaluate the effectiveness of our framework.
The results show that 
1) the effectiveness depends strongly on the nature of the task,
2) our proposed techniques greatly enhance effectiveness and robustness,
and 3) existing online fine-tuning methods are overall less effective than expected, calling for research to address it.

\end{description}

\section{Preliminaries}
\paragraph{Markov Decision Processes.}
We consider formalizing (online/offline) RL with (discrete-time,infinite horizon) Markov decision processes (MDPs), specified by
tuple $M=(\StateSpace, \ActionSpace, p, r, \gamma)$.
At each time step $h$, the agent observes the state $s_h\in\StateSpace$ and takes an action $a_h\in\ActionSpace$;
the next state $s_{h+1}$ is sampled according to probability distribution $p(\cdot|s_h, a_h)\in\Delta_{\StateSpace}$, while the 
reward function $r:\StateSpace\times\ActionSpace \to \R$ yields a scalar feedback $r_h:=r(s_h,a_h)$.
A policy $\pi=\{\pi(\cdot|s)\}_{s\in\StateSpace}\in\Pi$ consists of distributions $\pi(\cdot|s)\in\Delta_{\ActionSpace}$.
We use $\E_\pi[\cdot]$ to denote the expectation with respect to the random trajectory induced by $\pi$ in MDP $M$, that is, $(s_0, a_0, r_0, s_1, a_1, r_1, \ldots)$ where $s_0$ is a given initial state and $a_h\sim\pi(\cdot|s_h)$.
Given a trajectory, the discounted return is $G_h:=\sum_{h'=h}^\infty \gamma^{h'-h} r_{h'}$ with discount factor $\gamma\in[0,1)$.
For policy $\pi$, let $V^\pi:\StateSpace\to\R$ and $Q^\pi:\StateSpace\times\ActionSpace\to\R$ denote its state- and action-value functions, defined as 
$V^\pi(s):=\E_\pi[G_0|s_0=s]$ and 
$Q^\pi(s,a):=\E_\pi[G_0|s_0=s, a_0=a]$.
There always exists an optimal policy $\pi^* := \argmax_\pi V^\pi(s)$ that maximizes the state values for all $s\in\StateSpace$ simultaneously.
Let $V^*:=V^{\pi^*}$ and $Q^*:=Q^{\pi^*}$
denote the optimal value functions of $\pi^*$.

\paragraph{Online Reinforcement Learning.}
In the setting of online RL (from scratch), the agent is assumed to know state space $\StateSpace$, action space $\ActionSpace$, and discount factor $\gamma$ of $M$ but not transition probability $p$ or reward function $r$.
Without any other prior knowledge, the goal is to find a policy $\widehat{\pi}\in\Pi$ after sequentially taking a certain amount of actions and observing the resulting trajectories, such that $V^*-V^{\widehat{\pi}}$ is small.
In this paper, we consider MDPs with terminal states that mark the end of an episode, after which the agent is transitioned into a new state to start the next episode.
Formally, the information at online step $t$ is $(s_t,a_t,r_t,c_t)$, where $c_t\in\{0,1\}$ is the continuation flag such that
$s_{t+1}\sim p(\cdot|s_t,a_t)$ if $c_t=1$ (i.e., the episode continues) 
and $s_{t+1}\sim d_0\in\Delta_{\StateSpace}$ (i.e., the episode ends) where $d_0$ is the episode initial state distribution.
An online RL algorithm consists of a {\em behavior strategy} $\beta(\cdot)$ and a {\em recommendation strategy} $\widehat{\pi}(\cdot)$, where $\beta\left(s_t, s_{<t},a_{<t},r_{<t},c_{<t}\right)=:\beta_t\in\Delta_{\ActionSpace}$ determines the next action to take give the up-to-date information, i.e., $a_t\sim\beta_t$,
and $\widehat{\pi}(s_{< t},a_{< t},r_{< t},c_{< t})=:\widehat{\pi}_t\in\Pi$ to output a policy after taking exactly $t$ actions.
This definition is general enough to fit all online RL algorithms.
For example,
the Soft Actor-Critic (SAC) \citep{haarnoja2018soft_a,haarnoja2018soft_b} is developed as an online algorithm, which maintains a action-value function approximator $Q_
\theta$ 
and a policy $\pi_\theta$ 
that are jointly parameterized by $\theta$.
At online step $t$, SAC chooses action $a_t$ according to the current parameter $\theta_t$ as $a_t\sim\beta_t = \pi_{\theta_t}(\cdot|s_t)$.
Meanwhile, SAC maintains a replay buffer that stores (a subset of) transitions up to $t$ and updates parameter $\theta_t$ to $\theta_{t+1}$.
SAC's recommended strategy for evaluation is simply the deterministic version of the behavior policy, i.e., $\widehat{\pi}_t(s) = \argmax_{a} \pi_{\theta_t}(a|s)$.

The recommendation strategy is only relevant if the evaluation metric entails outputting a policy that is not necessarily the same as the ones used in the online steps.
In this paper, we focus on such an evaluation metric and note here that there exist other evaluation metrics, e.g., online regret \citep{auer2006logarithmic,auer2008near}, that do not require separately outputting a policy.

\paragraph{Offline Reinforcement Learning.}
In the setting offline RL, the goal is to find a good policy solely from a static dataset $\ReplayBuffer=\{(s_i,a_i,r_i,s'_i)\}$ obtained in $M$, where $r_i=r(s_i,a_i)$ and $s'_i\sim p(\cdot|s_i,a_i)$.
Thus, an offline RL algorithm is characterized by its extraction for a policy given the dataset, denoted as $\widehat{\pi}_{\rm off}(\ReplayBuffer)\in\Pi$.
It is believed that the core challenge of offline RL is how to estimate and incorporate values for actions that are not in the dataset.
Implicit Q-learning (IQL) \citep{kostrikov2022offline} is an offline RL algorithm that address the challenge by 
only performing value maximization for actions in the dataset,
which parameterizes state- and action-value functions
and learns their parameters 
with certain expectile value 
to approximate $V^*$ and $Q^*$ considering only actions in the dataset, without referring to actions not in the dataset for which the values are erroneous.
A parameterized policy 
is extracted by minimizing 
the advantage-weighted regression objective
that assigns larger weights to higher-advantage state-action pairs.

\section{Related Work}\label{sec:related_work}

Before introducing our framework, we review existing work on connecting offline and online RL, which has studied two problem settings:
\begin{itemize}[wide, 
labelwidth=!, labelindent=0pt, 
]
    \item \textit{Offline-to-Online RL} assumes access to a pre-collected offline dataset or a policy before online learning begins. Prior works in this direction can be categorized into two groups: Policy-Value Adaptation, which focuses on adapting the offline-learned Q-function~\citep{nakamoto2023calql} or policy~\citep{PEX} for smoother online fine-tuning, and Offline Data Utilization (ODU), which explores different ways to incorporate offline datasets into online learning. Within the ODU group, some methods \citep{zheng2023adaptive, lee2022offline, mark2022fine} first train a policy using offline data before refining it online, while others~\citep{niu2022trust, song2022hybrid, nair2020awac, vecerik2017leveraging} incorporate offline datasets without requiring policy pre-training. While these methods enhance online RL, they all require a pre-collected offline dataset, with some relying on high-quality offline data for effective performance.
    
    \item \textit{Deployment-Efficient RL} trains a policy offline from scratch with a focus on minimizing the number of changes in the data-collection policy, a.k.a, deployment efficiency. In this setting, after each data collection, an RL algorithm trains a policy offline. Two model-based offline RL methods~\citep{matsushima2020deployment, su2021musbo} are proposed to optimize deployment efficiency.
\end{itemize}

Our work differs from these prior works by sticking with the {\em tabula rasa} online RL setting, where the agent learns from scratch, constructs its own dataset for offline learning, and incorporates offline RL subroutines within the online learning process. We leave an extended review of the related work in Appendix \ref{asec:Extended Related Work}.

\section{Offline RL As Online Subroutines}
This paper focuses on online RL
where the agent uses the behavior strategy of an online RL algorithm to sequentially take a {\rm budget} of $T$ actions online, resulting in a {\em process} of $\T_T:=(s_t,a_t,r_t,c_t)_{t=0}^{T-1}$;
it then computes the final policy using the recommendation strategy as $\widehat{\pi}_T = \widehat{\pi}(\T_T)$ to maximize $J(\widehat{\pi}_T)$ where $J(\pi):=\E_{s\sim d_0}[V^\pi(s)]$.

\paragraph{SAC$@T$.}
Standard practice of online RL is to adopt a ``purely'' online RL algorithm like SAC, which 
often updates its learnable parameters in an incremental manner.
We use \SAC$@T$ to label the process of running SAC for exactly $T$ online steps and subsequently recommending the corresponding policy,
which is used as the (purely online) baseline to compare with our proposed online algorithms that incorporate offline RL subroutines as described next.

\begin{figure*}[htb]
\begin{center}
\includegraphics[width=.779\linewidth]{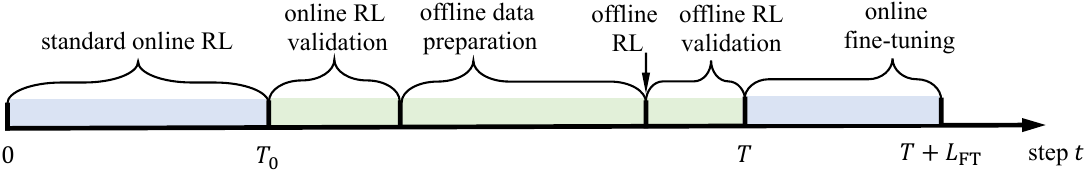}
\caption{Schematic of our incorporation of offline RL and online fine-tuning as online subroutines.}
\label{fig:schematic}
\end{center}
\end{figure*}

\subsection{Offline RL As Recommendation Strategies}
\label{sec:Offline RL as Recommendation Strategies}
Since offline RL holds the promise of extracting from some arbitrary dataset policies that attain values higher than the dataset's average returns, it is therefore natural to employ offline RL as the recommend strategy of an online algorithm.
In this Section \ref{sec:Offline RL as Recommendation Strategies}, we propose several such online algorithms that output their final policy $\widehat{\pi}_T$ by calling an offline RL algorithm, where the differences between these online algorithms lie in how the offline dataset is generated.

\paragraph{IQL on SAC.}
As the most straightforward way, we first run SAC until the entire budget $T$ is used up, and then perform IQL by repurposing the entire online interactions as the offline dataset.
This final output policy can therefore be denoted as $\widehat{\pi}^{\rm IQL}(\ReplayBuffer(\T_T^{\rm SAC}))$, where $\widehat{\pi}^{\rm IQL}(\cdot)$ is the IQL algorithm itself and $\ReplayBuffer(\T_T^{\rm SAC})$ is the dataset constructed by reformatting the $T$ online steps generated by (the behavior strategy of) SAC.
We label this process as \IQL(\SAC).

\paragraph{SAC + IQL.}
The performance of policies extracted by offline RL is known to be sensitive to the given dataset, specifically its
size \citep{cheng2024look}
and quality 
\citep{schweighofer2022dataset}.
In the naive process of \IQL(\SAC), the dataset for offline RL comes directly from the original SAC, which is oblivious to the fact that IQL could have benefited from a more carefully constructed dataset.
Motivated by this, we  propose a simple alternative process where the offline dataset is separately collected using the best policy over a set of periodically evaluated ones from SAC.
Formally, this process takes the following procedures:
\begin{enumerate}[label=(\roman*), wide, labelwidth=!, labelindent=0pt, itemsep=-.5pt, topsep = 0pt]
\item  
Run  SAC that terminates early at online step $T_0<T$, during which a set of $K$ policies $\{\widehat{\pi}_{t_k}\}_{k=1}^{K}$ are stored periodically at online steps $(t_1, \ldots, t_K=T_0)$ with $\widehat{\pi}_{t_k}$ obtained from SAC's recommendation strategy;

\item \label{SAC+IQL step:dataset_preparation}
Using another $T_1-T_0$ steps, perform Monte Carol estimation on the policy values, $\widehat{J}(\widehat{\pi}_{t_k})\approx J(\widehat{\pi}_{t_k})$.
Then, choose the policy of the best estimate, $\pi_{\ReplayBuffer}:=\argmax_{\widehat{\pi}_{t_k}}\widehat{J}(\widehat{\pi}_{t_k})$,  and use it to collect a dataset $\ReplayBuffer$ with the remaining $T-T_1$ steps;

\item
Output the final policy by  IQL, i.e., $\widehat{\pi}^{\rm IQL}(\ReplayBuffer)$.
\end{enumerate}
This process is labeled \SAC+\IQL, which by design takes exactly $T$ online steps.
Because of phase (ii)'s ``best pooling'' operation, referred to as {\em online RL validation}, the collected dataset $\ReplayBuffer$ for IQL is expected to be of higher quality than that in \IQL(\SAC) and therefore potentially leads to better offline learned policy. 


\subsection{Offline RL Followed by Online Fine-Tuning}
\label{sec:Offline RL Followed by Online Fine-Tuning}
Section \ref{sec:Offline RL as Recommendation Strategies} employs an offline RL algorithm as the online subroutine of recommendation strategy.
Existing offline RL works have also considered the setting of {\em online fine-tuning} where, after finishing the offline learning on a given dataset, the agent starts online interactions and continues its learning.
We here explore incorporating such a fine-tuning phase following the offline RL.
Specifically, we consider the following alternatives for the fine-tuning phase.

\paragraph{IQL-based fine-tuning.}
The original IQL paper \citep{kostrikov2022offline} uses the exactly same IQL training objective during for their fine-tuning, where both both offline and newly collected online data are used.
As a natural option, we also adopt such a fine-tuning (FT) phase, labeled as \FT[\IQL], which follows the offline IQL phase of \IQL(\SAC) or \SAC+\IQL{}.
Formally, after running offline IQL on dataset $\ReplayBuffer$ to obtain learned policy $\pi_\theta$,
perform another $L_\FT$ online steps with actions sampled by policy $\pi_{\theta_l}$ for $l=1,\ldots,L_\FT$, with $\theta_1=\theta$ and $\theta_{l>1}$ obtained by performing IQL updates with transitions sampled from the combination of dataset $\ReplayBuffer$ and the first $l-1$ online fine-tuning transitions.
Output the final policy as $\pi_{\theta_{L_\FT}}$.

\paragraph{PEX.}
IQL is not an algorithm specifically designed for fine-tuning.
Recent work has offered algorithms specific to fine-tuning that are claim to be more efficient.
We here consider PEX \citep{PEX}, which adopts the idea of policy expansion for fine-tuning and is shown to be more effective than IQL on the D4RL benchmark.
Specifically, after running IQL on dataset $\ReplayBuffer$ to obtain learned policy $\pi_\beta$,
expand it into a policy set $\Pi=[\pi_\beta, \pi_\theta]$, where $\pi_\theta$ is a randomly initialized, online-learnable policy.
Perform $L_\FT$ online steps with actions sampled by the policy set $\Pi_l=\{\pi_\beta, \pi_{\theta_l}\}$ for $l=1,\ldots,L_\FT$, with $\Pi_l=\Pi$ and $\Pi_{l>1}$ obtained by performing IQL updates on $\pi_{\theta_l}$ using a dataset equally drawn from $\ReplayBuffer$ and the online replay buffer, i.e., the first $l-1$ online fine-tuning transitions. $\Pi_l$ derives a policy by selecting from itself a candidate policy with a higher Q value and then using it to sample an action. The final policy is derived from $\Pi_{L_\FT}=\{\pi_\beta, \pi_{\theta_{L_\FT}}\}$.

\paragraph{Cal-QL.}
Prior work has observed that many algorithms experience an performance drop at the beginning of fine-tuning, with the performance only being recovered/improved later.
\citet{nakamoto2023calql} analyzed this phenomenon and developed their algorithm, Cal-QL, which they claim learns a calibrated initialization during offline RL such that the following fine-tuning is faster and more effective.
We therefore include Cal-QL in our experiments, by replacing IQL with Cal-QL in both the offline RL phase and the online fine-tuning phase, resulting in two processes respectively labeled as \CalQL(\SAC)+\FT[\CalQL] and \SAC+\CalQL+\FT[\CalQL].

\subsection{Offline RL Validation}
\label{sec:Offline RL Validation}
In the strictly offline RL setting, the agent cannot obtain more data other than the given dataset. Therefore, if the agent needs to validate its learned policy for, for example, early stopping its gradient update to prevent overfitting, it has to rely on the given dataset itself, in a similar fashion of the validation data split in supervised setting.
While offline RL in principle can perform such validation, standard offline algorithms rarely do it for simplicity, i.e.,
they often just perform a fixed number of gradient updates before outputting the final policy without validation.

In our setting, however, the agent has the luxury of performing interactions to validate the offline RL procedure. 
Note that these interactions count towards the total budget.
We implement offline RL validation in the similar manner as online validation's ``best-pooling'' operation: we periodically pause the offline RL gradient update to evaluate the policy at that point and do best-pooling after all gradient updates.

\paragraph{Summary.}
Figure \ref{fig:schematic} is a schematic showing all the procedures introduced. An entire online RL process consists of a subset of them, if not all.
As illustrated, an online process will begin with a standard online RL phase, possibly followed by an offline RL phase that optionally makes additional online interactions for data preparation and/or validation, possibly ending with fine-tuning.

\section{Experiments}
\label{sec:experiments}
\paragraph{Environments.}
Our experiments are conducted over twelve diverse environments from the following four groups, which have been used as popular benchmarks for offline RL, e.g., D4RL \citep{fu2020d4rl}, varying significantly in reward types and task goals.
The {\it Franka Kitchen} group manipulates kitchen appliances with sparse rewards, requiring precise coordination of complex tasks;
{\it Adroit Hand} focuses on object manipulation with a robotic hand, featuring dense rewards and demanding fine motor control. 
{\it MuJoCo} group simulates locomotion tasks where robots must coordinate their limbs to achieve smooth movement, with dense rewards guiding their progress. 
{\it Point Maze} involves navigating a ball through 2D mazes with sparse rewards.
More details can be found in Appendix \ref{asec:env_under_study}.

\paragraph{Hyperparameters.}
Our experiments investigate our proposed online processes with a total budget of 1500K steps, including $T$ steps for incorporating offline RL and $L_\FT$ steps for fine-tuning.
To fully define an online process, we need to determine the hyperparameters of $T$, $L_\FT$, and the budgets for all the procedures including standard online RL ($T_0$) and online/offline validation (cf. Figure \ref{fig:schematic}).
To avoid excessive tuning, we do not tune these hyperparameters and instead choose their values in a reasonable manner, as detailed in Appendix \ref{asec:selection_of_online_RL_processes}.
For example, for \SAC+\IQL+\FT{}, we set $T=$1100K and $L_\FT=$400K, with $T_0=$800K, 200K for dataset collection, and 50K each for offline and online validation.
We will describe our hyperparameters choices when presenting the results.  


\paragraph{Baselines and normalization.}
For reference, we also run SAC as a purely online baseline for $T+L_\FT = $1500K steps.
We report the mean and standard error of the normalized return across 10 seeds.
We normalize a policy's return such that, over the 10 seeds, a random policy has a score of 0 and \SAC$@$1500K has a score of 100. 


We present our results to respectively analyze 
the effectiveness of using offline RL as the recommendation strategy compared to a purely online baseline (Section \ref{sec:Results of Offline RL as Recommendation Strategy})
and the effectiveness of the online fine-tuning (Section \ref{sec:Effectiveness of Online Fine-Tuning}).

\begin{figure*}[htb]
    \centering
    \includegraphics[width=\textwidth]{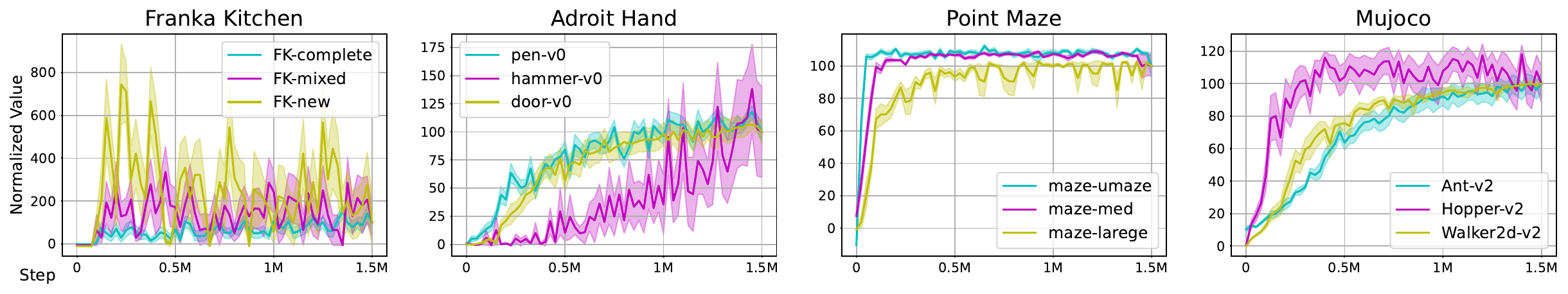}
    \caption{Learning curves of the purely online RL process of \SAC$@$1500K on all environments.}
    \label{fig:learning_curves_purely_online_all_envs}
\end{figure*}

\subsection{Results of Offline RL as Recommendation Strategy}
\label{sec:Results of Offline RL as Recommendation Strategy}
\begin{table}[htb]
\centering
\fontsize{7}{10}\selectfont

\begin{tabular}{l||c|c||c|c}
  & \multicolumn{2}{c||}{$T=$1050K}&
  \multicolumn{2}{c}{$T=$1100K} \\ \hline

Environment    & \SAC$@T$  & \IQL(\SAC)   & \SAC$@T${}  & \SAC+\IQL \\ \hline

FK-complete  & 72.2 ± 30.2   & \textbf{146.3}±34.1     & 50.0±27.0   & \textbf{148.1}±37.3 \\ 
FK-mixed   & 114.3±82.5  & \textbf{407.1}±114.1 & 207.1±113.7 & \textbf{553.6}±107.7  \\ 
FK-new     & 211.1±148.1 & \textbf{850.0}±126.8 & 100.0±111.1 & \textbf{950.0}±100.1\\ \hline

pen-v0    & 108.1±8.9   & \textbf{128.1}±1.3   & 106.6±6.2 & \textbf{116.9}±2.3 \\ 
hammer-v0  & 38.8±19.6   & \textbf{180.3±}38.8   & 103.4±40.5  & \textbf{128.3}±32.1 \\ 
door-v0    & 100.3±10.2 & \textbf{101.6}±11.4 & \textbf{101.7}±7.8   & 87.8±14.8 \\ \hline

Ant-v2    & \textbf{94.5}±6.1    & 92.5±4.9  & \textbf{92.8}±5.7    & 92.5±5.2  \\ 
Hopper-v2   & \textbf{115.2}±7.4   & 108.4±7.3        & 107.4±10.1  & \textbf{121.0}±2.1   \\ 
Walker2d-v2     & \textbf{96.0}±2.7    & 82.8±5.6    & \textbf{94.5}±2.7  & 90.6±1.7     \\ \hline

maze-umaze        & 109.3±1.1   & \textbf{120.6}±0.8     & 106.3±1.5   & \textbf{120.2}±1.0 \\ 
maze-med      & 105.8±0.7   & \textbf{111.6}±0.4      & 105.8±1.0   & \textbf{111.6}±0.3    \\ 
maze-large       & \textbf{99.3}±0.7    & 96.6±8.6       & 101.1±1.5   & \textbf{105.2}±1.3    

\end{tabular}
\caption{Normalized values of \IQL(\SAC) and \SAC+\IQL. In \textbf{bold} is the higher mean between one of the two processes and its purely online reference with the same budget $T$.}
\label{tab:offlineRL_main}
\end{table}

We examine the two online processes introduced in Section \ref{sec:Offline RL as Recommendation Strategies}, \IQL(\SAC) and \SAC+\IQL, both using IQL to recommend the final policy after $T$ steps, with the difference being how the offline dataset is prepared.
For \IQL(\SAC), we use 1000K steps for SAC and 50K for offline validation, resulting in $T=$1050K.
For \SAC+\IQL, we use 800K steps for SAC, 200K to collect the dataset, and 50K each for online and offline validation, resulting in $T=$1100K.
Table \ref{tab:offlineRL_main} presents the normalized policy values, with \SAC$@T$ evaluated at corresponding budgets for reference.
We make the following observations and analyses.

\paragraph{Environment-dependent effectiveness.}
As shown in Table \ref{tab:offlineRL_main}, for most environments, using offline IQL as the recommendation strategy improves the final policy over the purely online \SAC$@T$.
A key observation is that the effectiveness of the offline phase is highly associated with the environment nature.
Specifically, the improvement is most significant in the three Franka Kitchen (FK) environments and the Adroit Hand environment of hammer-v0, where SAC turns out to be ineffective and/or unstable as shown in Figure \ref{fig:learning_curves_purely_online_all_envs}.
For the other environments, SAC is stable and effective to yield a near-convergent policies after $T$ steps, and therefore an offline RL recommendation strategy does not significantly improve them.

We hypothesize FK's sparse rewards is a key factor that makes SAC difficult to learn.
To test this hypothesis, we here modify the the MuJoCo environments to introduce sparse rewards.
We accumulate the original rewards over a number of time steps, called the reward accumulation steps (ras), and assign the accumulated reward as the reward every ras steps and a reward of zero at other time steps.
This reward shaping enables adjustable reward sparsity (by adjusting ras) while ensuring the optimal policy has the same value.
Table \ref{tab:mujoco_sparse_rewards} presents the results on several MuJoCo sparse reward variants that support our hypothesis.
Here, SAC yields policies with significantly lower values, which are significantly improved by the offline IQL. 
\begin{table}[htb]
\centering
\fontsize{7.7}{10}\selectfont

\begin{tabular}{l||c|c||c|c}
& \multicolumn{2}{c||}{$T=$1050K}&
  \multicolumn{2}{c}{$T=$1100K} \\ \hline
Environment   & \SAC$@T$ & \IQL(\SAC)  & \SAC$@T$ & \SAC+\IQL \\ \hline
Ant-v2-ras20   & 14.0±5.9   & \textbf{15.0}±6.0   & 14.5±6.2          & \textbf{15.0}±4.3  \\
Hopper-v2-ras30  & 36.3±11.8  & \textbf{55.9}±13.3 & 38.0±11.6         & \textbf{51.0}±12.0 \\
Walker2d-v2-ras30 & 21.1±4.3  & \textbf{43.3}±7.1  & 20.5±4.8          & \textbf{24.3}±5.5\\
\end{tabular}

\caption{Normalized values in the MuJoCo environments with sparse rewards.
In \textbf{bold} is the higher mean between one of the two processes and its purely online reference with the same budget $T$.
}
\label{tab:mujoco_sparse_rewards}
\end{table}

\paragraph{Ablation: online/offline validation.}
Both \IQL(\SAC) and \SAC+\IQL{} use offline RL validation (cf. Section \ref{sec:Offline RL Validation}) in their IQL, with \SAC+\IQL{} additionally performs online RL validation (its step \ref{SAC+IQL step:dataset_preparation}) for its dataset preparation.
We perform an ablation study here to assess the importance of these validations.
For fair comparison, we evaluate the variants with and without validations that all use $T=$1000K steps:
\IQL(\SAC) without validation uses all 1000K steps in its SAC online learning phase;
\IQL(\SAC) with validation uses 950K in its SAC and 50K for validation;
\SAC+\IQL{} without validation uses 800K steps for SAC and 200K to collect the offline dataset with SAC's last updated policy;
\SAC+\IQL{} with validation uses 700K for SAC, 200K for data collection, and 50K each for online and offline validation.

Table \ref{tab:table_ablation_offline_RL_validation} presents the results, which clearly show the importance of online/offline RL validation for any environment, especially for those where SAC is ineffective/unstable. 
\begin{table}[htb]
\centering
\fontsize{7}{10}\selectfont

\begin{tabular}{l||cc||cc}
 &
  \multicolumn{2}{c||}{\IQL(\SAC), $T=$1000K} &
  \multicolumn{2}{c}{\SAC+\IQL, $T=$1000K} \\ \hline
Environment &
  \multicolumn{1}{c|}{\begin{tabular}[c]{@{}c@{}}w/o val.\end{tabular}} &
  \begin{tabular}[c]{@{}c@{}}w/ val.\end{tabular} &
  \multicolumn{1}{c|}{\begin{tabular}[c]{@{}c@{}}w/o val.\end{tabular}} &
  \begin{tabular}[c]{@{}c@{}} w/ val.\end{tabular} \\ \hline
FK-complete   & \multicolumn{1}{c|}{83.3±25.7}   & \textbf{150.9}±26.2  & \multicolumn{1}{c|}{89.8±31.2}          & \textbf{158.3}±19.7       \\
FK-mixed  & \multicolumn{1}{c|}{232.1±94.5}  & \textbf{453.6}±77.2  & \multicolumn{1}{c|}{189.3±95.9}         & \textbf{521.4}±111.7      \\
FK-new  & \multicolumn{1}{c|}{383.3±135.2} & \textbf{750.0}±127.8 & \multicolumn{1}{c|}{211.1±148.1}        & \textbf{966.7}±92.3        \\ \hline
pen-v0    & \multicolumn{1}{c|}{96.9±6.0}    & \textbf{130.2}±2.1   & \multicolumn{1}{c|}{96.4±4.7}           & \textbf{118.9}±2.2       \\
hammer-v0    & \multicolumn{1}{c|}{118.7±34.7}  & \textbf{137.8}±32.4  & \multicolumn{1}{c|}{22.7±9.2}           & \textbf{158.2}±24.8      \\
door-v0   & \multicolumn{1}{c|}{91.9±12.8}   & \textbf{94.6}±11.0  & \multicolumn{1}{c|}{85.6±10.1}          & \textbf{93.3}±10.9         \\ \hline
Ant-v2    & \multicolumn{1}{c|}{91.9±6.0}    & \textbf{94.7}±3.8    & \multicolumn{1}{c|}{81.5±7.3}           & \textbf{91.0}±5.7          \\
Hopper-v2   & \multicolumn{1}{c|}{80.7±11.1}   & \textbf{111.7}±8.6   & \multicolumn{1}{c|}{103.8±9.9}          & \textbf{121.2}±1.7        \\
Walker2d-v2   & \multicolumn{1}{c|}{76.6±9.7}    & \textbf{88.1}±4.8    & \multicolumn{1}{c|}{87.2±3.3}           & \textbf{88.3}±2.0          \\ \hline
maze-umaze     & \multicolumn{1}{c|}{105.9±2.0}   & \textbf{120.5}±0.9  & \multicolumn{1}{c|}{101.2±5.5}          & \textbf{121.0}±0.9        \\
maze-med  & \multicolumn{1}{c|}{104.1±1.6}   & \textbf{111.0}±0.6   & \multicolumn{1}{c|}{105.2±1.5}          & \textbf{111.7}±0.5         \\
maze-large  & \multicolumn{1}{c|}{72.5±11.9}   & \textbf{103.3}±1.9   & \multicolumn{1}{c|}{87.7±8.3}           & \textbf{104.5}±1.5        
\end{tabular}%
\caption{Ablation of offline and online RL validation.
In \textbf{bold} is the higher mean between the two variants of with and without validation.
}
\label{tab:table_ablation_offline_RL_validation}
\end{table}

\begin{table*}
\centering
\fontsize{7.7}{10}\selectfont
\begin{tabular}{l||c|c|c|c|c|c}
                     & \multicolumn{2}{c|}{{\IQL(\SAC)}}  & \multicolumn{2}{c|}{{\SAC+\IQL}} & {\CalQL(\SAC)}                  & {\SAC+\CalQL}                  \\ 
\cline{1-5}
{Environment} & {+\FT[\IQL]} & {+\FT[\PEX]} & {+\FT[\IQL]} & {+\FT[\PEX]} & {+\FT[\CalQL]}                 & {+\FT[\CalQL]}                 \\ 
\hline
FK-complete  & $\downarrow$ 55.6±21.3          & $\uparrow$ \textbf{157.4}±26.7         & $\downarrow$118.5±32.9         & $\uparrow$153.7±35.4         & 114.8±28.4 
 $\downarrow$ 41.2±30.2   & 141.7±39.8 $\downarrow$ 25.3±18.1    \\
FK-mixed     & $\downarrow$185.7±90.4         & $\rightarrow$ 407.1±116.1        &  $\downarrow$500±90.6           & $\uparrow$\textbf{557.1}±90.6         & 210.7±96.3 $\downarrow$ 204.8±107.9 & 589.3±157.5 $\downarrow$ 55.8±49.4   \\
FK-new       & $\downarrow$566.7±134.4        & $\uparrow$900.0±107.1        & $\uparrow$\textbf{977.8}±115.9        & $\downarrow$922.2±103.7        & 5.6±8.5 $\uparrow$  127.5±111.2    & 927.8±117.8 $\downarrow$ 68.8±64.5  \\ 
\hline
pen-v0               & $\downarrow$\textbf{101.0}±5.2          & $\downarrow$92.1±7.1           & $\downarrow$72.4±6.3           & $\downarrow$85.4±8.4           & 21.1±3.3 $\uparrow$ 25.2±4.6      & 21.3±3.5 $\downarrow$15.3±5.8       \\
hammer-v0            & $\downarrow$\textbf{98.6}±34.2          & $\downarrow$94.7±31.0          & $\downarrow$82.1±33.2          & $\downarrow$63.8±28.2          & 18.9±8.0 $\downarrow$ -1.4±0.9      & 1.6±0.7 $\uparrow$ 7.5±6.6        \\
door-v0              & $\downarrow$82.3±10.8          & $\downarrow$\textbf{91.8}±10.9          & $\downarrow$70.6±12.7          & $\downarrow$77.0±13.5          & 0.1±0.0 $\uparrow$ 50.9±10.4      & 0.1±0.0 $\uparrow$ 24.8±12.5      \\ 
\hline
Ant-v2               & $\downarrow$\textbf{85.1}±5.7           & $\downarrow$79.6±6.4           & $\downarrow$63.5±6.0           & $\downarrow$83.6±5.8           & 18.9±0.0 $\uparrow$ 80.5±8.1      & 18.9±0.0 $\uparrow$ 76.4±7.8      \\
Hopper-v2            & $\downarrow$102.5±11.0         & $\downarrow$77.4±10.0          & $\downarrow$\textbf{120.1}±2.9          & $\downarrow$72.7±9.1           & 54.2±8.7 $\downarrow$ 49.7±7.8       & 27.8±5.3 $\uparrow$ 58.4±9.7      \\
Walker2d-v2          & $\uparrow$\textbf{88.1}±2.3           &$\downarrow$76.5±5.3           & $\downarrow$66.0±7.8           & $\downarrow$63.4±15.3          & 12.7±1.8 $\uparrow$ 68.3±8.1      & 11.9±3.2 $\uparrow$ 41.8±10.7     \\ 
\hline
maze-umaze      & $\downarrow$106.8±1.4          & $\downarrow$95.6±6.0           & $\downarrow$\textbf{109.0}±1.5          & $\downarrow$94.4±3.9           & 113.8±2.6 $\downarrow$ 106.9±1.5    & 102.0±4.9 $\downarrow$ 72.7±5.5     \\
maze-med     & $\downarrow$106.3±1.1          & $\downarrow$104.1±0.9          &$\downarrow$\textbf{106.7}±0.8          & $\downarrow$99.5±3.2           & 88.1±6.4 $\uparrow$ 102.9±2.0     & 102.1±4.8 $\downarrow$ 75.8±8.2     \\
maze-large      &$\uparrow$\textbf{591.1}±9.2          & $\downarrow$93.9±2.2           & $\downarrow$94.2±2.4           & $\downarrow$71.3±11.8          & 86.1±5.7 $\downarrow$ 85.1±3.3      & 55.1±9.5 $\uparrow$ 58.7±6.1     
\end{tabular}%

\caption{Normalized values by the end of the fine-tuning phase (before arrow), as well as before fine-tuning for the CalQL-based processes (after arrow).
Note that the values before fine-tuning are reported in Table \ref{tab:offlineRL_main} for the IQL/PEX-based processes.
The arrows ($\uparrow$, $\downarrow$) indicate whether the mean increases or decreases by the end of the fine-tuning.
In \textbf{bold} is the highest mean among the six online learning processes {\em by the end} of the fine-tuning.
}
\label{tab:fine_tuning_main}
\end{table*}

\paragraph{Comparing the data preparation strategies.}
We next analyze the difference between \IQL(\SAC) and \SAC+\IQL, which differ in how the offline dataset is prepared.
For fair comparison, we analyze Table \ref{tab:table_ablation_offline_RL_validation} with $T=$1000K for both processes (w/ val.) to make the following observations: 
\begin{itemize}[wide, labelwidth=!, labelindent=0pt, itemsep=-.5pt, topsep = -.5pt]
\item 
The difference in policy values is again environment-dependent, \SAC+\IQL{} significantly outperforms \IQL(\SAC) in the same four environments (three FK ones and hammer-v0) where SAC struggles, while the two processes are equally effective in MuJoCo and Point Maze. 
This suggests that it is most beneficial to allocate steps dedicated to dataset collection when learning with a purely online algorithm is ineffective/unstable.

\item
\IQL(\SAC) is better in pen-v0 and door-v0, where SAC makes steady progress yet does not converge by $T$=1000K (cf. Figure \ref{fig:learning_curves_purely_online_all_envs}).
Therefore, the difference is likely due to the fact that \IQL(\SAC) can leverage the diversity in the replay buffer, while \SAC+\IQL{} only exploits the information in a prematurely trained policy.

\end{itemize}

\subsection{Effectiveness of Online Fine-Tuning}
\label{sec:Effectiveness of Online Fine-Tuning}

We next analyze the online-fine tuning phase, which succeeds the offline RL phase and concludes the entire online process.
As introduced in Section \ref{sec:Offline RL Followed by Online Fine-Tuning}, we evaluate six such processes:
the first four use \IQL(\SAC) or \SAC+\IQL{} for the offline RL phase as described in Section \ref{sec:Results of Offline RL as Recommendation Strategy}, followed by IQL-based (\FT[\IQL]) or PEX-based (\FT[\PEX]) fine-tuning;
the other two replace IQL with Cal-QL in both the offline (\CalQL(\SAC) or \SAC+\CalQL{} and fine-tuning (\FT[\CalQL]) phases.

Table \ref{tab:fine_tuning_main} presents the normalized values by the end of fine-tuning, where all six processes have consumed $T+L_{\FT}=$1500K steps.
Perhaps surprisingly, the results show that fine-tuning is not as ineffective as expected:
\begin{itemize}[wide, labelwidth=!, labelindent=0pt, itemsep=-.5pt, topsep = -.5pt]
\item
IQL-based fine-tuning \FT[\IQL] actually degrades the policy's mean value in almost all cases, with the only exception being \IQL(\SAC)+\FT[\IQL] in Walker2d-v2. Yet, in this case, the final policy is still worse than that from the purely online \SAC$@$1500K (normalized to 100).

\item
PEX-based fine-tuning \FT[\PEX] degrades the policy's mean value in all environments except those in the Franka Kitchen group.

\item
Cal-QL-based fine-tuning \FT[\CalQL] improves the policy's mean value in more cases.
However, the resulting policy by the end of fine-tuning is worse than the IQL/PEX-based counterparts in all environments, implying that Cal-QL is ineffective in the offline RL phase. 
\end{itemize}

We below provide there analyses on various failure modes that explain the ineffective of the fine-tuning.

\paragraph{Analysis 1: IQL-based fine-tuning in MuJoCo.}
Table \ref{tab:fine_tuning_main} shows that IQL-based fine-tuning decreases the policy performance in most of the environments, which contradicts prior studies \citep{PEX,nakamoto2023calql} showing that IQL improves policies initialized from offline learning with D4RL datasets in the MuJoCo environments.
We identify here that a key difference between our setup and these studies is the budget for fine-tuning:
We limit the total budget to 1500K steps, resulting in fine-tuning steps of 450K and 400K steps for fine-tuning in \IQL(\SAC)+\FT[\IQL] and \SAC+\IQL+\FT[\IQL], respectively, while the prior studies allocate 1000K steps for fine-tuning.

\begin{figure}[htb]
\begin{center}
\includegraphics[width=.9\columnwidth]{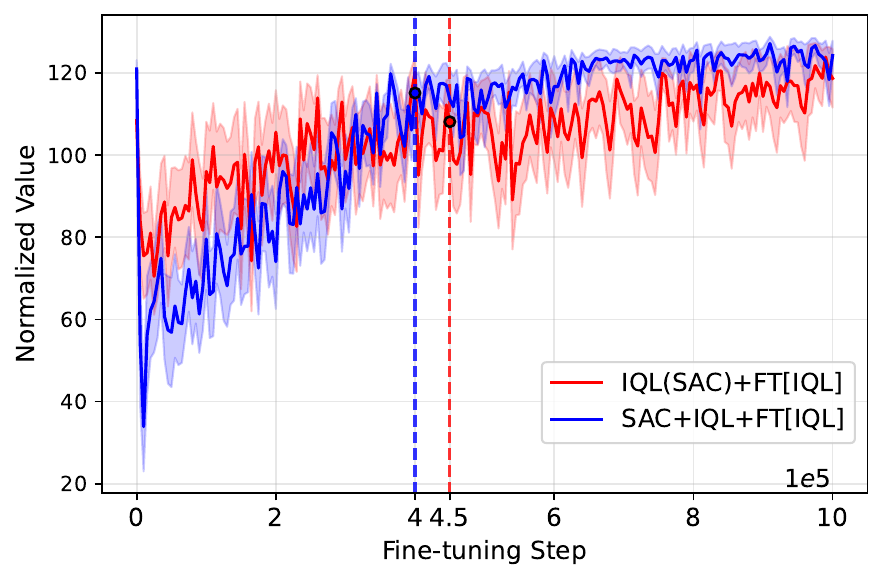}
\caption{Extending IQL-based fine-tuning to 1000K steps in Hopper-v2.
The dashed lines indicate the original fine-tuning budgets as reported in Table \ref{tab:fine_tuning_main}.}
\label{fig:IQL_FT1M_Hopper}
\end{center}
\end{figure}

Figure \ref{fig:IQL_FT1M_Hopper} shows the learning curves where we extend the number of steps for our IQL-based fine-tuning to 1000K in Hopper-v2.
The policy performance drastically drops at the beginning, recovers later, and finally is improved over the offline initialization by the end of 1000K fine-tuning steps.
This suggests that, while improvement is eventually possible, IQL-based fine-tuning is budget-inefficient.

\paragraph{Analysis 2: Cal-QL's offline learning in MuJoCo.}
Cal-QL was proposed to fix the issue of initial performance drop in pre-existing fine-tuning algorithms such as IQL.
Indeed, Table \ref{tab:fine_tuning_main} shows that Cal-QL's fine-tuning increases the policy performance from the offline initialization in a number of environments.
However, Cal-QL's fine-tuned policies are worse than those from IQL/PEX, because Cal-QL's offline learning is less effective.
Taking Hopper-v2 as an example, the offline RL of \CalQL(\SAC) achieves a mean normalized value of 54.2 (cf. Table \ref{tab:fine_tuning_main}), while \IQL(\SAC) achieves 108.4 (cf. Table \ref{tab:offlineRL_main}) and the original Cal-QL work \citep{nakamoto2024cal} achieves 103.0 from the counterpart D4RL dataset of hopper-medium-v2.
As we use the same Cal-QL code as in \citet{nakamoto2024cal}, the performance gap must come from differences in the offline datasets.

\begin{table}[htb]
\centering
\resizebox{\columnwidth}{!}{%
\begin{tabular}{l||c|c|c||c}
    Dataset &
  \begin{tabular}[c]{@{}c@{}}Size\end{tabular} &
  \begin{tabular}[c]{@{}c@{}}Avg. Return\end{tabular} &
  RPSV &
  \begin{tabular}[c]{@{}l@{}}\CalQL \end{tabular} \\ \hline
\CalQL(\SAC)            & 1000K & 49.6  & 1353  & 54.2±8.7     \\
\SAC+\CalQL             & 200K  & 122.3 & 8     & 27.8±5.3     \\
\SAC+\CalQL             & 1000K & 103.9 & 379     & 72.5±15.4     \\ \hline
hopper-medium-v2 & 1000K & 50.6  & 92    & 103.0 \\
hopper-expert-v2 & 1000K & 125.7 & 11    & 66.6
\end{tabular}%
}
\caption{
Statistics of the offline datasets collected in two of our Cal-QL online processes and two D4RL datasets in Hopper-v2.
The values of Cal-QL-optimized policies on the D4RL datasets are re-normalized with our scale based on the means reported in the Cal-QL paper, where the corresponding standard errors are not reported.}
\label{tab:hopper_dataset_stats}
\end{table}
\begin{figure}[tb]
\begin{center}
\includegraphics[width=.917\columnwidth]{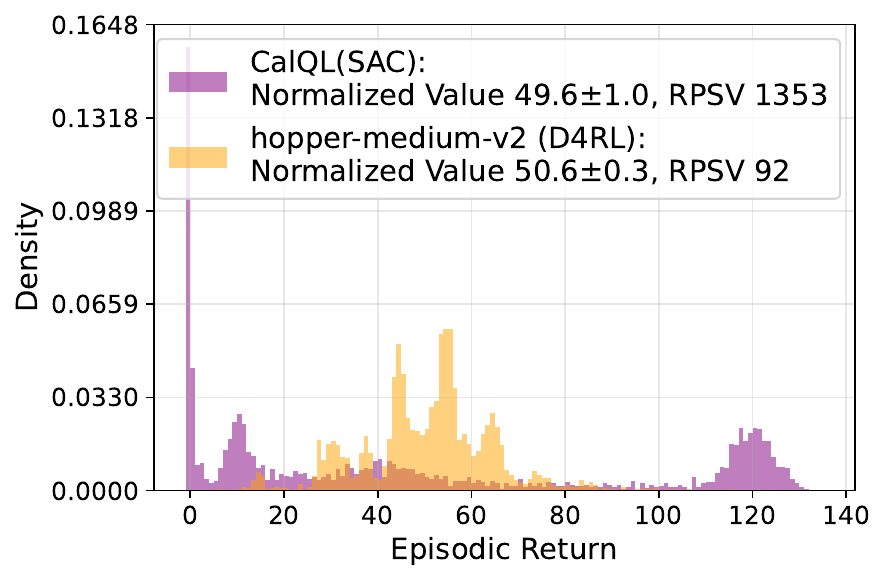}
\caption{Distributions of episodic returns (normalized) in 
\CalQL(\SAC)'s dataset and D4RL's hopper-medium-v2.}

\label{fig:hopper_dataset_dist}
\end{center}
\end{figure}

Table \ref{tab:hopper_dataset_stats} summarizes statistics of the relevant datasets. 
We provide the following observations and insights:
\begin{itemize}[wide, labelwidth=!, labelindent=0pt, itemsep=-.5pt, topsep = -.5pt]
\item 
\CalQL(\SAC)'s dataset has comparable average episodic return when compared to D4RL's hopper-medium-v2.
However, they differ significantly in episodic return's distribution as shown in Figure \ref{fig:hopper_dataset_dist}.
\CalQL(\SAC)'s dataset disperses away from the average episodic return toward a small number of high-return episodes, whereas hopper-medium-v2 remains more concentrated around its average. This difference is quantified by Return Positive-Side Variance (RPSV) proposed in \citet{hong2023harnessing}, which measures the variance of returns on the positive side of the mean.
As a reasonable explanation for the performance gap, offline RL algorithms such as Cal-QL fail to utilize high-return episodes in a high-RPSV dataset due to these algorithms' pessimistic and conservative nature.

\item 
For \SAC+\CalQL's dataset, the primary distinction from its D4RL counterpart, hopper-expert-v2, is its smaller size (200K vs. 1000K), which we hypothesize is the reason why \SAC+\CalQL's offline learning is ineffective. 
To verify this, we here collect a larger dataset of size 1000K using the same online validated policy and apply Cal-QL for offline learning. The result, as presented in the third row of Table \ref{tab:hopper_dataset_stats}, shows that increasing the dataset size significantly improves performance.
This indicates that the effectiveness of Cal-QL's offline learning strongly depends on the dataset size.
\end{itemize}

\paragraph{Analysis 3: Cal-QL's fine-tuning in Franka Kitchen.}
As another failure mode, Cal-QL fails to improve and even degrades the offline-learned policy during fine-tuning in FK-complete and FK-mixed when using our online-collected datasets (cf. Table \ref{tab:fine_tuning_main}). In contrast, the original Cal-QL paper \citep{nakamoto2024cal} reports policy improvement in these environments when using the corresponding D4RL datasets.

For FK-complete, a closer examination of the experiment setups reveals several key differences between ours and that in the Cal-QL paper, as shown in Table \ref{tab:experiment_setting_difference_CalQL_offline_Kitchen}.
Among these differences, we focus on the different in reward design, hypothesizing that the accumulated reward provides richer learning signals than the binary reward, thereby facilitating Cal-QL's fine-tuning. Specifically, this could lead to either an improvement in the offline initialization or a reduction in the performance drop observed during fine-tuning.
\begin{table}[tb]
\centering
\resizebox{\columnwidth}{!}{%
\begin{tabular}{c||cc|c}
                                                                         & \multicolumn{2}{c|}{Our  Setup} & Cal-QL's  Setup \\ \hline
\multicolumn{1}{c||}{\begin{tabular}[c]{@{}c@{}}Reward\end{tabular}}  &
  \multicolumn{2}{l|}{\begin{tabular}[c]{@{}l@{}}Binary Reward:\\ 1 if a subgoal is achieved\\ at the step; Otherwise, 0.\end{tabular}} &
  \multicolumn{1}{l}{\begin{tabular}[c]{@{}l@{}}Accumulated Reward:\\ the number of achieved \\ subgoals by the step.\end{tabular}} \\ \hline
\begin{tabular}[c]{@{}c@{}}Max Episode\\ Length\end{tabular}             & \multicolumn{2}{c|}{280}                  & 1000                      \\ \hline
\begin{tabular}[c]{@{}c@{}}Offline Dataset\\ Size\end{tabular} &
  \multicolumn{1}{c|}{\begin{tabular}[c]{@{}c@{}}\CalQL(\SAC) \\ Size of 1M\end{tabular}} &
  \begin{tabular}[c]{@{}c@{}}\SAC+\CalQL \\ Size of 200K\end{tabular} &
  \begin{tabular}[c]{@{}c@{}}D4RL (Human Demo) \\ Size of 3679\end{tabular} \\ \hline
\begin{tabular}[c]{@{}c@{}}Offline Dataset\\ Average Value\end{tabular} & \multicolumn{1}{c|}{69.44}    & 124.68    & 676.27                    \\ \hline
\begin{tabular}[c]{@{}c@{}}Fine-tuning\\ Steps\end{tabular}              & \multicolumn{1}{c|}{450K}     & 400K      & 1250K                    
\end{tabular}%
}
\caption{Comparing our Cal-QL's offline learning setup with the original \citep{nakamoto2023calql} in FK-complete.}
\label{tab:experiment_setting_difference_CalQL_offline_Kitchen}
\end{table}

To test this hypothesis, we here conduct an experiment in FK-complete where the binary reward is replaced with the accumulated reward during both offline learning and online-fine-tuning phases, while keeping using the original binary reward to for policy evaluation.
The normalized values of the offline-optimized policy and the fine-tuned policy in this new experiment are presented in Table \ref{tab:CalQL_binary_and_accumulated_rewards_in_kitchen}. 
Here, for both types of rewards, we also report the change ratio before and after fine-tuning.
The results show that the switch to accumulated rewards leads to better fine-tuning performance for Cal-QL. Specifically, Cal-QL exhibited either a lower value drop ratio in \CalQL(\SAC)+\FT[\CalQL] or a value gain in the fine-tuned policy in \SAC+\CalQL+\FT[\CalQL].

\begin{table}[htb]
\centering
\resizebox{\columnwidth}{!}{%
\begin{tabular}{c||c|c}
    &
  \begin{tabular}[c]{@{}c@{}}\CalQL(\SAC)\\ +\FT[\CalQL]\end{tabular} &
  \begin{tabular}[c]{@{}c@{}}\SAC+\CalQL\\ +\FT[\CalQL]\end{tabular} \\ \hline

\begin{tabular}[c]{@{}c@{}}Binary \end{tabular} & 114.8±28.4 $\downarrow$ 41.2±30.2 & 141.7±39.8 $\downarrow$ 25.3±18.1 \\ 
\begin{tabular}[c]{@{}c@{}}Reward\end{tabular} & -64.1\%                & -82.1\%               \\ \hline
\begin{tabular}[c]{@{}c@{}}Accumulated \end{tabular} &
  86.7±26.4 $\downarrow$ 50±22.8 &
  51.7±22.7 $\uparrow$ 56.9±34.9 \\ 
\begin{tabular}[c]{@{}c@{}}Reward\end{tabular} & -42.3\%                & 9.7\%                
\end{tabular}%
}
\caption{
Offline learning and fine-tuning results in FK-complete using binary and accumulated rewards.
}
\label{tab:CalQL_binary_and_accumulated_rewards_in_kitchen}
\end{table}

\section{Conclusion} \label{sec:conclusion}

We have formalized a general framework to incorporate offline RL, as well as the following fine-tuning, as subroutines of the tabula rasa online RL process.
To evaluate the effectiveness of the framework, we have conducted a systematic and extensive empirical study that compares the performance of incorporating recent offline RL and fine-tuning algorithms against the purely online RL baseline.

Our study has found that repurposing offline RL algorithms as online RL's recommendation strategy can significantly improve the policy, especially with proper validation and for environments where purely online RL is unstable/ineffective.
However, the evaluated fine-tuning methods are largely ineffective for further improving the offline-optimized policy, although they have been designed and tested for their effectiveness with benchmark offline datasets.
Our analyses have provided insights on several typical failure modes of the fine-tuning methods, showing that their performance can overly rely on the fine-tuning budget, the dataset's features, and/or the task nature.

\bibliographystyle{abbrvnat}
\bibliography{references}

\newpage
\onecolumn
\appendix

\section{Extended Related Work}
\label{asec:Extended Related Work}
This section expands upon the related work briefly mentioned in the main paper, providing a detailed exploration of the key contributions and limitations in the hybrid Offline-to-Online area. Additionally, it discusses the Deployment-Efficient Reinforcement Learning Problem, which, like ours, starts policy learning from scratch with repeated online data collection and offline policy updates, highlighting the differences from our problem setting. This extended review underscores the progress and challenges in hybrid offline-online RL, setting the stage for the novel contributions of this study, which aims to explore how offline RL techniques can be effectively integrated into a tabula rasa online RL framework.

\subsection{Offline-to-Online}
In the Offline-to-Online problem setting, an offline dataset or pre-optimized policy is available before the online RL process begins. Recent advancements have significantly improved learning efficiency and policy performance in this area. This body of work can be categorized into the following groups.

\begin{itemize}
    \item \textbf{Using Offline Learning to Establish a Strong Initial Policy for Online Reinforcement Learning}: \citep{nakamoto2024cal} presents Cal-QL, a method designed to enable rapid online fine-tuning of a policy learned from an offline dataset. It mitigates the initial performance drop during fine-tuning by ensuring that the offline-learned Q-function is constrained to be above a reference policy's value, keeping the Q-values within a reasonable scale. \citep{PEX} introduce a policy expansion method designed to facilitate a smoother transition from offline to online RL. Their approach strategically balances conservatism with exploration by incrementally expanding the policy learned from offline data during the online phase. This method leverages the stability of offline learning while adapting to new information acquired online. \citep{zheng2023adaptive} propose an adaptive framework that dynamically integrates offline data into the online learning process. Their approach enhances policy performance by selectively incorporating relevant offline experiences, which guide the online exploration process. This adaptive integration aims to mitigate the common issue of distributional shift between offline data and the online environment. While effective, the method relies heavily on the quality and relevance of the offline data, which may not always be available. \citep{lee2022offline} address the challenge of sample efficiency by introducing a balanced replay mechanism coupled with a pessimistic Q-ensemble. This combination is designed to tackle distributional shifts by maintaining a conservative estimate of the Q-values during online learning. The balanced replay mechanism ensures that both offline and online experiences contribute effectively to the learning process. Despite its strengths, this approach presupposes a high-quality offline dataset, limiting its applicability in fully online or data-scarce environments. \citep{mark2022fine} proposes an exploration technique that enables stable and efficient online fine-tuning while maintaining the same training objective as in offline RL. The method selects optimistic actions, which are estimated to be better than those suggested by the training policy, generating exploratory data to improve the policy. Unlike previous approaches, this work addresses the more general scenario where the offline data may be of low or mixed quality.
    \item \textbf{Leveraging Offline Datasets to Accelerate Online Reinforcement Learning}: \citep{niu2022trust} introduce a dynamics-aware policy evaluation scheme within the H2O framework, which addresses the challenges of learning effective reinforcement learning policies with imperfect simulators. H2O adaptively adjusts the Q-values of simulated state-action pairs by penalizing or boosting them based on the estimated dynamics gap between real and simulated environments. This approach mitigates bias from simulator imperfections, effectively bridging the sim-to-real gap. The authors assume that leveraging a limited offline dataset with good coverage of real-world dynamics, despite the simulator's imperfections, can enhance learning. \citep{song2022hybrid} presents Hybrid Q-Learning (Hy-Q), a novel algorithm designed for Hybrid Reinforcement Learning (Hybrid RL), which integrates offline datasets with online interactions. Hy-Q builds on classical Q-learning to enhance both statistical and computational efficiency, particularly when offline data supports a high-quality policy and the environment exhibits bounded bilinear rank. \citep{nair2020awac} propose the Advantage Weighted Actor Critic (AWAC) algorithm, which leverages offline datasets to accelerate online reinforcement learning. AWAC combines the stability of offline learning with the flexibility of online exploration by reweighting actions based on their offline advantage, thus prioritizing promising actions during policy updates. Unlike other methods, AWAC avoids overly conservative updates by not explicitly modeling the behavior policy, leading to more balanced policy improvements. \citep{vecerik2017leveraging} introduces a method that effectively combines expert demonstrations with actual interactions to tackle the challenge of sparse rewards in robotic environments. Both demonstrations and real-time experiences are stored in a replay buffer, with the sampling ratio between the two automatically adjusted using a prioritized replay mechanism. This approach allows the agent to efficiently learn from successful examples while still exploring the environment, significantly improving the learning process in complex robotics tasks where sparse rewards make exploration difficult.
\end{itemize}

While the aforementioned studies contribute valuable insights into the integration of offline and online RL, they share a common limitation: the assumption that an offline dataset and/or a reference policy exists before the commencement of online learning. This assumption restricts the applicability of these methods in scenarios where the agent must learn from scratch (tabula rasa) without any pre-existing data or policies. In contrast, the present study focuses on scenarios where the agent starts with no prior knowledge, and explores how offline RL can be incorporated as a subroutine within an entirely online learning process. This approach is particularly relevant for environments where acquiring an initial offline dataset with sufficient coverage is impractical.

\subsection{Deployment-efficient Reinforcement Learning}
Deployment-efficient RL problem is motivated by the fact that, in many real-world applications of RL, such as health~\citep{murphy2001marginal}, education~\citep{mandel2014offline} and dialog agents~\citep{jaques2019way}, implementing a new data-collection policy may incur various costs and pose certain risks. So, it is ideal to minimize the number of changes in the data-collection policy during learning, a.k.a, deployment efficiency. The approaches~\citep{matsushima2020deployment, su2021musbo} for deployment-efficient RL problem start policy learning from scratch and iteratively collect data online while updating the policy offline.

Our problem setting shares similarities with deployment-efficient RL in that both start from scratch and interleave online-offline phases. However, there are key differences: (1) While deployment-efficient RL problem prioritizes minimizing policy deployments, our setting allows for a more flexible integration of offline and online phases, where the online phase involves a full-fledged online RL process. (2) Our offline phase can include additional data collection, using the output policy from the preceding online phase to prepare the dataset for offline RL, which is not typically considered in deployment-efficient RL. (3) The sequence of phases in our approach is not rigidly structured and can conclude in either the online or offline phase, providing greater adaptability to different environments and learning scenarios. These distinctions highlight the broader applicability and flexibility of our method in optimizing RL performance across varied contexts.


\newpage\clearpage
\section{Environments Under Study}\label{asec:env_under_study}
11 of 12 studied environments are adopted from D4RL~\citep{fu2020d4rl}; the newly added one is a Franka Kitchen environment, kitchen-new-v0, where the agent aims to complete the list of tasks, [’kettle’, ’light switch’, ’microwave’, ’hinge cabinet']. The goal of each rest environment are listed in the Table~\ref{tab:goals_of_12_envs}. And each environment's characteristics, including state shape, action shape, reward type, are listed in Table~\ref{tab:characteristics_of_12_envs}.

\begin{table*}[ht]
\centering
\fontsize{7}{10}\selectfont
\begin{tabular}{c||c||c}
Environment Name    & Task Goal                                                                                                          & Environment Group               \\ \hline
kitchen-new-v0      & Complete the list of tasks {[}'kettle', 'light switch', 'microwave',  'hinge cabinet'{]}                           & \multirow{3}{*}{Franka Kitchen} \\ 
kitchen-mixed-v0    & Complete the list of tasks {[}'microwave', 'kettle', 'light switch', 'slide cabinet'{]}                           &                                  \\ 
kitchen-complete-v0 & Complete the list of tasks {[}'microwave', 'kettle', 'bottom burner', 'light switch'{]}                           &                                  \\ \hline
door-v0  & The task to be completed consists on undoing the latch and swing the door open.                                & \multirow{3}{*}{Adroit Hand}    \\ 
pen-v0       & The task to be completed consists on repositioning the blue pen to match the orientation of the green target. &                                 \\ 
hammer-v0    & The task to be completed consists on picking up a hammer with and drive a nail into a board.                  &                                 \\ \hline
Ant-v2              & The goal is to coordinate the four legs to move in the forward (right) direction                              & \multirow{3}{*}{Mujoco}         \\ 
Hopper-v2           & The goal is to make hops that move in the forward (right) direction                                           &                                 \\ 
Walker2d-v2         & The goal is to make coordinate both sets of feet, legs, and thighs to move in the forward (right) direction   &                                 \\ \hline
maze2d-umaze-v1     & Move the green ball to the goal location in a u-shape maze                                                                     & \multirow{3}{*}{Point Maze}     \\ 
maze2d-medium-v1    & Move the green ball to the goal location in a medium-size maze                                                                     &                                 \\ 
maze2d-large-v1     & Move the green ball to the goal location in a large-size maze                                                                     &                                 \\ 
\end{tabular}%
\caption{The goals of 12 environments used in our investigation.}
\label{tab:goals_of_12_envs}
\end{table*}

\begin{table}[htb]
\centering
\begin{tabular}{c||c|c|c|c}
Environment Name    & State Space & Action Space & Reward Type & Environment Group               \\ \hline
kitchen-new-v0      &             &              &             & \multirow{3}{*}{Franka Kitchen} \\ 
kitchen-mixed-v0    & (59, )      & (9, )        &Sparse       &                                 \\ 
kitchen-complete-v0 &             &              &             &                                 \\ \hline

door-v0             & (39, )      & (28, )       &             & \multirow{3}{*}{Adroit Hand}    \\ 
pen-v0              & (45, )      & (24, )       &Dense        &                                 \\ 
hammer-v0           & (46, )      & (26, )       &             &                                 \\ \hline

Ant-v2              & (27, )      & (8, )        &             & \multirow{3}{*}{Mujoco}         \\ 
Hopper-v2           & (11, )      & (3, )        &Dense        &                                 \\ 
Walker2d-v2         & (17, )      & (6, )        &             &                                 \\ \hline

maze2d-umaze-v1     &             &              &             & \multirow{3}{*}{Point Maze}     \\ 
maze2d-medium-v1    & (8, )       & (2, )        &Sparse       &                                 \\ 
maze2d-large-v1     &             &              &             &                                 \\ 
\end{tabular}%
\caption{The characteristics of each of the 12 environments used in our investigation.}
\label{tab:characteristics_of_12_envs}
\end{table}

\newpage\clearpage
\section{Configuration Selection for Studied Online RL Processes}\label{asec:selection_of_online_RL_processes}
An online RL process within our proposed framework can have multiple configurable parameters, including the initial online stop step, data collection budget for the \textit{NewData} strategy, fine-tuning budget, online/offline RL validation budget, number of evaluations per validation, and the number of episodes per evaluation. The \textit{NewData} strategy refers to collect a new dataset using the output policy of the initial online phase of an online RL process, e.g., \SAC+\IQL{}. Also, we define the \textit{Buffer} strategy as directly using the online replay of the initial online phase of an online RL process, e.g., \IQL(\SAC).  In this study, we primarily focused on two key parameters: the \textit{initial online stop step} and the \textit{online/offline RL validation budget}, before finalizing the online RL processes for our investigation.

\subsection{Initial Online Stop Step}
We explored two values for the \textit{initial online stop step} for each offline dataset preparation strategy, \textit{Buffer} and \textit{NewData}. Since using a replay buffer of 1000K transitions as the offline-learning dataset is a common baseline in offline RL literature, the first \textit{initial online stop step} selected for the \textit{Buffer} strategy was 1000K, resulting in the online RL process \IQL(SAC$@\text{1000K}$){}.

To create an online RL process with a concluding offline phase using the \textit{NewData} strategy, where the agent takes the same number of steps as in \IQL(SAC$@\text{1000K}$){} during training, we designed the process \SAC($@\text{800K}$)+\IQL{}. Here, the agent stops the initial online phase at 800K steps and then uses the resulting policy to collect a dataset of 200K transitions for IQL during offline learning.

Additionally, we explored another set of online RL processes, \IQL(SAC$@\text{600K}$){} and \SAC($@\text{400K}$)+\IQL{}, where the initial online stop steps were set to 600K and 400K, respectively. Note that each online or offline RL validation consumes 50K steps.

As shown in Table~\ref{tab:offline_phases_with_different_online_stop_step}, in many environments, both of \IQL(SAC$@\text{600K}$){} and \SAC($@\text{400K}$)+\IQL{} online RL processes outperform their purely online RL counterpart processes, \SAC$@\text{650K}${} with a budget of 650K steps and \SAC$@\text{700K}${} with a budget of 700K steps, respectively. Similarly, as seen in Table~\ref{tab:offlineRL_main}, \IQL(SAC$@\text{1000K}$){} and \SAC($@\text{800K}$)+\IQL{} beat their purely online RL counterpart processes, \SAC$@\text{1050K}${} and \SAC$@\text{1100K}${}, respectively. Given that both sets of online RL processes demonstrate the benefit of integrating an offline phase, and since \IQL(SAC$@\text{1000K}$){} aligns with a common baseline in offline RL literature, we decided to continue our studies using \IQL(SAC$@\text{1000K}$){} and \SAC($@\text{800K}$)+\IQL{}.

\begin{table}[htb]
\centering
\begin{tabular}{l||c||c|c||c|c}
Environment         & \begin{tabular}[c]{@{}c@{}}\SAC$@\text{1500K}$\\ $T0=\text{1500K}$\end{tabular} & \begin{tabular}[c]{@{}c@{}}\SAC$@\text{650K}$\\ $T0=\text{650K}$\end{tabular} & \begin{tabular}[c]{@{}c@{}}\IQL(\SAC$@\text{600K}$)\\ $T=\text{650K}$\end{tabular} & \begin{tabular}[c]{@{}c@{}}\SAC$@\text{700K}$\\ $T0=\text{700K}$\end{tabular} & \begin{tabular}[c]{@{}c@{}}\SAC$@\text{400K}$+\IQL\\ $T=\text{700K}$\end{tabular} \\ \hline
kitchen-complete-v0 & 100.0±35.1  & 38.9±24.3   & \textbf{159.3±26.2}  & 70.4±29.5  & \textbf{137.0±23.9}  \\ 
kitchen-mixed-v0    & 100.0±71.6  & 71.4±71.0   & \textbf{532.1±105.5} & 150.0±84.4 & \textbf{471.4±120.8} \\ 
kitchen-new-v0      & 100.0±111.1 & 177.8±111.2 & \textbf{866.7±83.6}  & 144.4±94.0 & \textbf{1111.1±81.5} \\ \hline
pen-v0              & 100.0±8.6   & 92.4±6.9    & \textbf{119.4±2.7}   & 86.1±5.4   & \textbf{104.5±2.6}   \\ 
hammer-v0           & 100.0±40.3  & 21.5±19.9   & \textbf{78.3±29.0}   & 19.9±13.2  & \textbf{40.0±16.1}   \\ 
door-v0             & 100.0±10.6  & 81.7±10.2   & 81.6±10.7            & 92.8±10.2  & 79.5±8.0             \\ \hline
Ant-v2              & 100.0±6.1   & 78.4±5.0    & \textbf{87.2±4.2}    & 77.9±6.4   & 73.6±3.9             \\ 
Hopper-v2           & 100.0±11.5  & 105.5±9.8   & \textbf{112.8±6.1}   & 113.6±7.9  & \textbf{117.1±0.6}   \\ 
Walker2d-v2         & 100.0±2.1   & 87.5±2.8    & \textbf{87.8±1.8}    & 89.8±2.4   & 75.9±4.1             \\ \hline
maze2d-umaze-v1     & 100.0±8.1   & 112.2±1.3   & \textbf{120.9±0.8}   & 109.9±1.5  & \textbf{120.9±0.8}   \\ 
maze2d-medium-v1    & 100.0±6.3   & 107.9±0.8   & \textbf{111.3±0.5}   & 107.3±0.4  & \textbf{111.4±0.6}   \\ 
maze2d-large-v1     & 100.0±1.1   & 90.4±9.3    & \textbf{104.9±1.3}   & 100.4±1.0  & \textbf{105.8±1.6}   \\ 
\end{tabular}%
\caption{Normalized values of output policies from the purely online RL processes and the online RL processes with an concluding offline phase. \SAC$@\text{650K}${} with a budget of 650K steps and \SAC$@\text{700K}${} with a budget of 700K steps are included as baselines to compare with \IQL(\SAC$@\text{600K}$) and \SAC$@\text{400K}$+\IQL{} respectively. The normalized value of an offline-optimized policy is made bold to highlight when the offline phase outperforms the purely online RL process under the same budget.}
\label{tab:offline_phases_with_different_online_stop_step}
\end{table}

\subsection{RL Validation Budget}
Given the considerably higher standard error of the means  (as seen Table~\ref{tab:offlineRL_main})) for the scores of learned policies in the Franka Kitchen and Adroit Hand environments, compared to those in the Mujoco and Point Maze environments, we decided to set the number of episodes per evaluation to 20 for the Franka Kitchen and Adroit Hand environments, and to 5 for the Mujoco and Point Maze environments when working with the large budget of 1500K steps.
For each of the three RL validation budgets—20K, 50K, and 100K—the number of evaluations per validation in each environment is listed in Table~\ref{tab:num_evals_per_validation_by_validation_budget}. The number of evaluations per RL validation corresponds to the number of data points in the best pooling process, meaning that a larger number of evaluations increases the likelihood of extracting the best policy during training.
As seen in Table~\ref{tab:num_evals_per_validation_by_validation_budget},  with a validation budget of 20K, the best pooling process in the Franka Kitchen environments will only have three evaluations (data points) per validation, making the process highly unstable. Therefore, we focused on exploring the other two validation budgets.
Table~\ref{tab:impact_of_validation_budget} shows that the online RL process with a 100K validation budget outperforms the process with a 50K budget in 10 out of 12 environments. However, a significant improvement of more than 15\% is consistently seen in only three environments, while in most of the others, the improvement from doubling the validation budget is less than 2.5\%. Based on these observations, we decided to set the RL validation budget to 50K steps.

\begin{table}[htb]
\centering
\begin{tabular}{l|c|c|ccc}
\multicolumn{1}{c|}{\multirow{2}{*}{Environment}} &
  \multirow{2}{*}{\begin{tabular}[c]{@{}c@{}}Max Steps\\ Per Episode\end{tabular}} &
  \multirow{2}{*}{\begin{tabular}[c]{@{}c@{}}Number of\\ Episodes \\      Per Evaluation\end{tabular}} &
  \multicolumn{3}{c}{Number of Evaluations Per RL Validation} \\ \cline{4-6} 
\multicolumn{1}{c|}{} &
   &
   &
  \multicolumn{1}{c|}{\begin{tabular}[c]{@{}c@{}}Validation Budget \\ 20K\end{tabular}} &
  \multicolumn{1}{c|}{\begin{tabular}[c]{@{}c@{}}Validation Budget\\ 50K\end{tabular}} &
  \begin{tabular}[c]{@{}c@{}}Validation Budget\\ 100K\end{tabular} \\ \hline
kitchen-mixed-v0    & 280  & 20 & \multicolumn{1}{c|}{3}  & \multicolumn{1}{c|}{8}  & 17 \\ 
kitchen-complete-v0 & 280  & 20 & \multicolumn{1}{c|}{3}  & \multicolumn{1}{c|}{8}  & 17 \\ 
kitchen-new-v0      & 280  & 20 & \multicolumn{1}{c|}{3}  & \multicolumn{1}{c|}{8}  & 17 \\ \hline
door-v0             & 200  & 20 & \multicolumn{1}{c|}{5}  & \multicolumn{1}{c|}{10} & 25 \\ 
pen-v0              & 100  & 20 & \multicolumn{1}{c|}{10} & \multicolumn{1}{c|}{20} & 50 \\ 
hammer-v0           & 200  & 20 & \multicolumn{1}{c|}{5}  & \multicolumn{1}{c|}{10} & 25 \\ \hline
Ant-v2              & 1000 & 5  & \multicolumn{1}{c|}{4}  & \multicolumn{1}{c|}{10} & 20 \\ 
Hopper-v2           & 1000 & 5  & \multicolumn{1}{c|}{4}  & \multicolumn{1}{c|}{10} & 20 \\ 
Walker2d-v2         & 1000 & 5  & \multicolumn{1}{c|}{4}  & \multicolumn{1}{c|}{10} & 20 \\ \hline
maze2d-umaze-v1     & 300  & 5  & \multicolumn{1}{c|}{13} & \multicolumn{1}{c|}{33} & 66 \\ 
maze2d-medium-v1    & 600  & 5  & \multicolumn{1}{c|}{6}  & \multicolumn{1}{c|}{16} & 33 \\ 
maze2d-large-v1     & 800  & 5  & \multicolumn{1}{c|}{5}  & \multicolumn{1}{c|}{12} & 25 \\ 
\end{tabular}%
\caption{Different RL validation budget results into different number of evaluations per validation.}
\label{tab:num_evals_per_validation_by_validation_budget}
\end{table}

\begin{table}[htb]
\centering
\fontsize{8}{11}\selectfont
\begin{tabular}{l|c|cc|cc}
\multirow{2}{*}{Environment} &
  \multirow{2}{*}{\SAC$@\text{1000K}${}} &
  \multicolumn{2}{c|}{\IQL(\SAC$@\text{1000K}$){}} &
  \multicolumn{2}{c}{\SAC$@\text{800K}$+\IQL{}} \\ \cline{3-6} 
 &
   &
  \multicolumn{1}{c|}{Validation Budget 50K} & Validation Budget 100K &
  \multicolumn{1}{c|}{Validation Budget 50K} & Validation Budget 100K \\ \hline
kitchen-complete-v0 & 100.00±25.31 & \multicolumn{1}{c|}{140.00±22.40} & 174.55±20.61 & \multicolumn{1}{c|}{146.36±35.26} & 201.82±22.64 \\ 
kitchen-mixed-v0 & 100.00±40.72 & \multicolumn{1}{c|}{198.53±50.28} & 230.88±47.35 & \multicolumn{1}{c|}{235.29±40.88} & 298.53±33.26 \\ 
kitchen-new-v0   & 100.00±99.78 & \multicolumn{1}{c|}{1580.00±182.57} & 1540.00±187.50 & \multicolumn{1}{c|}{1580.00±207.10} & 1990.00±10.00 \\ \hline
pen-v0    & 100.00±7.04  & \multicolumn{1}{c|}{140.37±2.54}  & 143.25±2.31  & \multicolumn{1}{c|}{122.94±3.12}  & 132.08±2.57  \\ 
hammer-v0 & 100.00±74.96 & \multicolumn{1}{c|}{380.59±73.78} & 468.99±92.44 & \multicolumn{1}{c|}{326.92±78.11} & 388.62±60.52 \\ 
door-v0   & 100.00±10.41 & \multicolumn{1}{c|}{108.15±5.76}  & 110.45±4.48  & \multicolumn{1}{c|}{97.77±11.37}  & 107.15±6.57  \\ \hline
Ant-v2           & 100.00±6.01  & \multicolumn{1}{c|}{103.71±6.13}  & 103.13±6.10  & \multicolumn{1}{c|}{99.75±5.12}   & 102.30±5.19  \\ 
Hopper-v2        & 100.00±7.57  & \multicolumn{1}{c|}{95.97±7.92}   & 104.93±6.02  & \multicolumn{1}{c|}{114.92±1.23}  & 113.17±1.48  \\ 
Walker2d-v2      & 100.00±2.76  & \multicolumn{1}{c|}{94.66±5.02}   & 94.86±4.68   & \multicolumn{1}{c|}{96.74±1.67}   & 94.53±1.81   \\ \hline
maze2d-umaze-v1  & 100.00±1.69  & \multicolumn{1}{c|}{112.01±1.12}  & 113.57±0.88  & \multicolumn{1}{c|}{110.92±0.90}  & 112.54±0.88  \\ 
maze2d-medium-v1 & 100.00±0.34  & \multicolumn{1}{c|}{103.63±0.70}  & 104.72±0.50  & \multicolumn{1}{c|}{104.01±0.37}  & 106.83±1.07  \\ 
maze2d-large-v1  & 100.00±1.09  & \multicolumn{1}{c|}{95.15±8.31}   & 100.93±6.10  & \multicolumn{1}{c|}{106.32±1.62}  & 108.64±0.78  \\ 
\end{tabular}%
\caption{The impact of the RL validation budget on the performance of an offline phase that concludes an online RL process.}
\label{tab:impact_of_validation_budget}
\end{table}

\newpage\clearpage
\section{Experimental Setup and Resources}\label{asec:expr_setup_and_resource}

\subsection{Implementation and Training Configuration of RL algorithms}\label{asec:train_configs_sac_and_iql}

In the online RL process, SAC is used for the initial online phase, while the offline phase follows either IQL or CalQL. In the fine-tuning phase, IQL or PEX is used if the offline phase was based on IQL, whereas CalQL is used if the offline phase followed CalQL.
The implementations of SAC and IQL are adopted from JaxRL~\citep{jaxrl}. For PEX, we ported the original PyTorch implementation of IQL-based PEX from the PEX authors~\citep{PEX} to a JAX-based implementation for our study. For CalQL, we uses the implementation from the CalQL authors~\citep{nakamoto2023calql}.

For the training configurations of SAC and IQL, we adhere to the settings outlined in the original papers. The common hyperparameters for SAC and IQL are listed in Table~\ref{tab:sac_iql_training_config}.
For IQL, the expectile and inverse temperature parameters are set per environment group as follows: Mujoco uses 0.7 and 3.0; Franka Kitchen and Adroit Hand use 0.7 and 0.5; and Point Maze uses 0.9 and 10.0, respectively.
Since we study the IQL-based PEX, both policies in PEX’s policy set are learned using IQL, meaning that PEX follows the same training configuration as IQL.
During the fine-tuning, both PEX and CalQL adopt the symmetric sampling to prepare batches for model update by randomly sampling equal number of transitions from the offline dataset and the online replay buffer.

\begin{table}[htb]
\centering

\begin{tabular}{l|c|c|c}
\textbf{Parameter}         & \textbf{SAC Value} & \textbf{IQL Value}    & \textbf{CalQL Value} \\ \hline
Actor Learning Rate        & 3e-4               & 3e-4                  & 1e-4                 \\
Critic Learning Rate       & 3e-4               & 3e-4                  & 3e-4                 \\
Value Learning Rate        & -                  & 3e-4                  & -                    \\
Temperature Learning Rate  & 3e-4               & -                     & -                    \\
Discount Factor            & 0.99               & 0.99                  & 0.99                 \\
Tau                        & 0.005              & 0.005                 & 0.005                \\
Initial Temperature        & 1.0                & -                     & 1.0                  \\
Critic Reduction           & -                  & Min                   & Min                  \\
Ensemble Size of Critics   & -                  & 2                     & 2                    \\
Target Network Update      & Moving Average     & Moving Average        & Moving Average       \\
Online Evaluation Interval & 5000               & 5000                  & 5000                 \\
Evaluation Episodes        & 10                 & 10                    & 10                   \\
Expectile                  & -                  & Environment Dependent & -                    \\
Inverse Temperature        & -                  & Environment Dependent & -                    \\
Learning Start Step        & 10000              & -                     & -                    \\
Offline Gradient Steps     & -                  & 500K                  & 500K                 \\
Online UTD Ratio           & -                  & -                     & 1.0                  \\
Automatic Entropy Tuning   & -                  & -                     & True                 \\
Batch Size                 & 256                & 256                   & 256                 
\end{tabular}
\caption{SAC, IQL and CalQL Training Configurations}
\label{tab:sac_iql_training_config}
\end{table}

\subsection{Computing Resources}
We conducted experiments in Linux machines with CUDA 12.3 and Nvidia GPUs, including Nvidia V100 (32G graphic memory per GPU unit) and Nvidia A100 (48G graphic memory per GPU unit).

\newpage\clearpage
\section{Average Returns of Random Policy and References Policies}\label{asec:perf_random_and_online}
In this study, we normalize a policy's return by using the average returns of a random policy as a baseline score of 0 and the average returns of a reference policy as a baseline score of 100. For a large budget of 1500K steps, the reference policy is the one produced by the purely online RL process, \SAC$@\text{1500K}$. For a smaller budget of 120K steps, the reference policy is the one produced by the purely online RL process, \SAC$@\text{100K}$. The reason for selecting the policy from \SAC$@\text{100K}$ as the reference for the smaller budget, instead of using the policy from \SAC$@\text{120K}$, is that the latter performs worse than the random policy in the hammer-v0 environment. This selection ensures a fairer comparison by avoiding negative scores for better-performing policies

\begin{table}[htb]
\centering
\begin{tabular}{l||c|c|c} 
Environment         & Random Policy & \SAC$@\text{1500K}${} & \SAC$@\text{100K}${} \\ \hline
kitchen-complete-v0 & 0.0±0.0       & 0.6±0.2           & 0.0±0.0          \\
kitchen-mixed-v0    & 0.0±0.0       & 0.2±0.1           & 0.1±0.1          \\
kitchen-new-v0      & 0.0±0.0       & 0.1±0.1           & 0.1±0.1          \\ \hline
pen-v0       & 56.6±13.9     & 3165.3±267.1      & 482.4±143.1      \\
hammer-v0    & -274.9±0.4    & 5182.6±2197.1     & 50.2±369.4       \\
door-v0      & -56.7±0.1     & 2776.6±299.1      & 59.1±71.6        \\ \hline
Ant-v2              & -51.1±11.7    & 5500.9±337.9      & 886.5±49.7       \\
Hopper-v2           & 18.0±1.2      & 2793.3±320.1      & 1226.9±266.2     \\
Walker2d-v2         & 1.2±0.7       & 5218.3±108.0      & 648.9±77.1       \\ \hline
maze2d-umaze-v1     & 27.2±5.3      & 236.2±17.0        & 250.3±2.0        \\
maze2d-medium-v1    & 18.5±5.9      & 476.0±28.8        & 473.3±13.4       \\
maze2d-large-v1     & 9.4±4.9       & 647.4±6.7         & 438.8±39.0       \\ \hline
Ant-v2-ras20        & -58.5±8.0     & 1360.8±346.0      & -48.0±12.7       \\
Hopper-v2-ras30     & 18.4±1.8      & 1073.4±306.0      & 371.5±33.0       \\
Walker2d-v2-ras30   & 1.2±0.3       & 2157.7±432.4      & 341.2±28.5       \\
\end{tabular}%
\caption{Average return of the random policy and the policies via the purely online processes with a budget of 1500K steps and 100K steps. The mean and its standard error (SEM) are reported in the format of $\text{Mean±SEM}$}
\label{tab:ave_ret_random_policy_and_online_policy}
\end{table}

\newpage\clearpage
\section{Change Ratio of Normalized Value: Offline-Enhanced vs. Purely Online RL Processes}
To facilitate the discussion of the possible benefit of an offline phase, we calculate the value change ratio of the output policy from an offline phase that concludes an online RL process with respect to the output policy from an purely online RL process with the same budget. Here, \SAC$@\text{1050K}${} and \SAC$@\text{1100K}${} are the purely onlile RL counterpart processes for \IQL(\SAC$@\text{1000K}$){} and \SAC$@\text{800K}$+\IQL{}, respectively.
\begin{table}[htb]
\centering
\begin{tabular}{l||c|c}
Environment         & \IQL(\SAC$@\text{1000K}$){} & \SAC$@\text{800K}$+\IQL{} \\ \hline
kitchen-complete-v0 & 102.57\%                    & 196.20\%                    \\
kitchen-mixed-v0    & 256.21\%                    & 167.26\%                    \\
kitchen-new-v0      & 302.63\%                    & 850.00\%                    \\ \hline
pen-v0       & 18.51\%                     & 9.70\%                      \\
hammer-v0    & 365.25\%                    & 24.12\%                     \\
door-v0      & 1.33\%                      & -13.65\%                    \\ \hline
Ant-v2              & -2.11\%                     & -0.28\%                     \\
Hopper-v2           & -5.93\%                     & 12.70\%                     \\
Walker2d-v2         & -13.78\%                    & -4.13\%                     \\ \hline
maze2d-umaze-v1     & 10.33\%                     & 13.04\%                     \\
maze2d-medium-v1    & 5.46\%                      & 5.49\%                      \\
maze2d-large-v1     & -2.74\%                     & 4.07\%                      \\ \hline
Ant-v2-ras20        & 7.13\%                      & 3.42\%                      \\ 
Hopper-v2-ras30     & 54.16\%                     & 34.14\%                     \\ 
Walker2d-v2-ras30   & 104.98\%                    & 18.37\%                     \\
\end{tabular}%
\caption{The value change ratio of the output policy from an online RL process with an ending offline phase with respect to the output policy of a purely online RL process with the same budget.}
\label{tab:change_ratio_offline_vs_online}
\end{table}

\newpage\clearpage
\section{Mujoco Environment With Sparse Reward}
The purpose of creating the MuJoCo environments with sparse rewards, denoted as \textit{MuJoCo (Sparse)}, is to test the hypothesis that an offline procedure could enhance learning efficiency in environments with sparse rewards. In the original MuJoCo environments, which feature dense rewards, the offline procedures studied in this paper did not consistently show a benefit in improving policy learning.

The \textit{MuJoCo (Sparse)} environments are variants of the original MuJoCo environments, created by converting their dense rewards into sparse rewards while keeping all other aspects unchanged. This conversion is achieved by accumulating the rewards from the original environment over a specified number of steps, referred to as reward accumulation steps (RAS), and then assigning the accumulated reward as the environment's reward every $RAS$ steps. For all other steps, the environment returns a reward of zero. This method ensures that the agent receives the same total reward in both the original MuJoCo environment and its sparse variant, allowing us to focus on analyzing the learning challenges introduced by reward sparsity.

Each \textit{MuJoCo (Sparse)} environment is labeled by appending "-ras{RAS}" to the original environment's name, where RAS indicates the number of steps over which rewards are accumulated. For instance, applying 20 reward accumulation steps to Hopper-v2 results in the variant environment Hopper-v2-ras20. In this variant, rewards from Hopper-v2 are accumulated every 20 steps and then returned as the reward for the corresponding step, while all other steps yield a reward of zero.

\begin{figure}[h]
    \centering
    \subfigure[Mujoco environment with different sparse rewards]{
        \centering
        \includegraphics[width=0.71\textwidth]{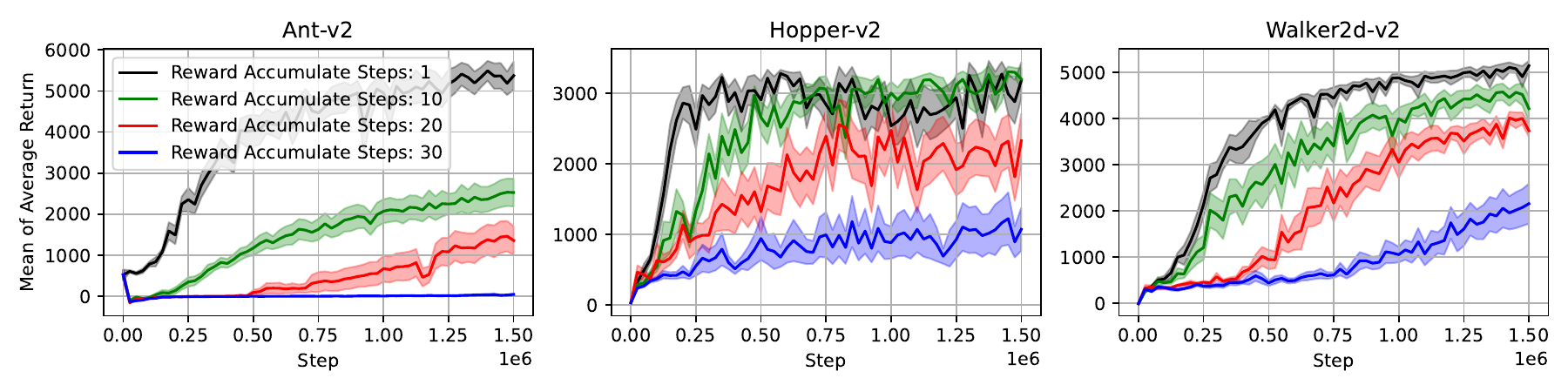}
        \label{fig:mujoco_sparse_envs_differernt_ras}
        \vspace{0.35cm}
    }
    \hfill
    \subfigure[Selected Mujoco (Sparse) environments for further study]{
        \centering
        \includegraphics[width=0.23\textwidth]{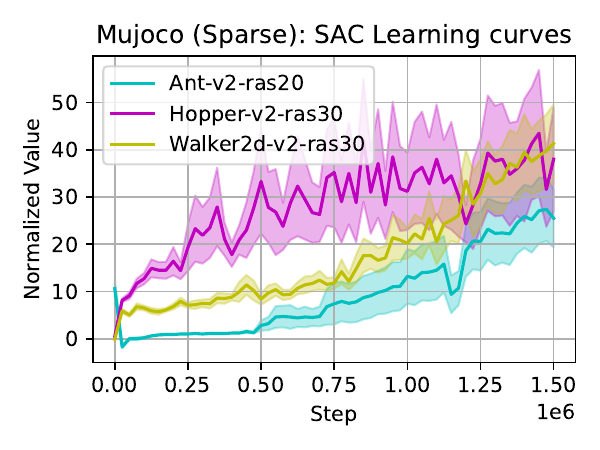}        \label{fig:learning_curves_purely_online_mujoco_ras} 
    }

    \caption{Learning curves of SAC in Mujoco (Sparse) environments with a budget of 1500K steps: (a) learning curves of SAC in Mujoco environments with different sparse rewards; (b) learning curves of SAC in selected Mujoco (Sparse) environments where scores are calculated by normalizing policies' returns using a randomly policy as a score of 0 and the policy by \SAC$@\text{1500K}$ in the original Mujoco environment as the score of 100.}
    \label{fig:mujoco_sparse_rewards_purelyonline_learn_curves}
\end{figure}

We explored three \textit{MuJoCo (Sparse)} environments, each with three different levels of reward sparsity, defined by RAS values of 10, 20, and 30. As shown in Figure~\ref{fig:mujoco_sparse_envs_differernt_ras}, in both Hopper-v2-ras30 and Walker2d-v2-ras30 environments, the learning pace of the purely online RL process was significantly slowed but still gradually progressed. In contrast, the Ant-v2-ras30 environment presented such a high learning difficulty that the agent failed to learn anything meaningful. However, in Ant-v2-ras20, while the learning difficulty was still substantially increased, the agent managed to make slow progress. Based on these observations, we selected Ant-v2-ras20, Hopper-v2-ras30, and Walker2d-v2-ras30 for further study in this paper. The learning curves of the purely online RL process using SAC in the three selected MuJoCo (Sparse) environments are shown in Figure~\ref{fig:learning_curves_purely_online_mujoco_ras}. These curves have been normalized, with a random policy set as a baseline score of 0, and the policy obtained by \SAC$@1500k$ in the original MuJoCo environment set as a score of 100.

\newpage\clearpage
\section{Behavior of the Offline Procedure in Limited Budget Scenarios}\label{asec:off_procedure_under_limited_budget}
The experiment discussed in the main paper assumes access to a substantial budget, specifically 1500K steps. However, in practical scenarios, the budget may be more limited.
Therefore, we further explore the potential benefits of an offline procedure that concludes an online RL process for policy learning under a smaller budget of 120K steps.
To do this, we investigate two additional online RL processes: \IQL(\SAC$@\text{100K}$){} and \SAC$@\text{80K}$+\IQL{}. 
It is important to note that the fine-tuning phase is excluded from this analysis, as our focus is on the scenario where an offline procedure is applied once to conclude the online RL process. 
In both \IQL(\SAC$@\text{100K}$){} and \SAC$@\text{80K}$+\IQL{}, each online and offline RL validation phase consumes 10K steps. 
In \SAC$@\text{80K}$+\IQL{}, a new dataset $\ReplayBuffer$ of 20K transitions is collected during the offline phase for IQL, whereas in \IQL(\SAC$@\text{100K}$){}, the existing replay buffer of 100K transitions is used as the offline training dataset for IQL.

As seen in Table~\ref{tab:small_budget_different_paths_in_each_env}, \IQL(\SAC$@\text{100K}$){} beats the corresponding purely online RL process, \SAC$@\text{110K}${}, in a large portion (8 out of 12) of the environments; while \SAC$@\text{80K}$+\IQL{} beats the corresponding purely online RL process, \SAC$@\text{120K}${}, in 4 out of 12 environments. Both processes underperform their purely online RL counterpart processes in all MuJoCo environments where purely online RL processes have a steady and sharp learning pace within the small budget window as shown in Figure~\ref{fig:learning_curves_purely_online_all_envs}. It is worth to note that \IQL(\SAC$@\text{100K}$){} beats \SAC$@\text{80K}$+\IQL{} with a big improvement in almost all environments except the environment, maze2d-uname-v1. In mazed2-umaze-v1, as shown in Table~\ref{tab:ave_ret_random_policy_and_online_policy}, the purely online RL process \SAC$@\text{80K}${} has achieved a near-optimal policy even with a small budget. This allows the best pooling after the online RL validation in \SAC$@\text{80K}$+\IQL{} to result in a very good (or near-optimal) policy. Consequently, a relatively better dataset can be collected for the offline procedure compared to \IQL(\SAC$@\text{100K}$){}. However, in other environments, the underperformance of \SAC$@\text{80K}$+\IQL{} is due to the limited budget, where the online phase has either not yet begun learning a good policy or has only just started. As a result, a good policy is difficult to retrieve through best pooling after the online RL validation, leading to the collection of relatively poor trajectories. Therefore, when considering applying an offline phase to a purely online phase under a limited budget of steps, the \textit{Buffer} strategy is preferred over the \textit{NewData} strategy.

\begin{table*}[htb]
\centering
\begin{tabular}{l||c||c|c||c|c}
Environment & \SAC$@\text{100K}${} & \SAC$@\text{110K}${} & \IQL(\SAC$@\text{100K}$){} & \SAC$@\text{120K}${} & \SAC$@\text{80K}$+\IQL{} \\ \hline
kitchen-complete-v0 & 100.00±150.00 & 250.00±300.00 & \textbf{800.00±316.67} & 950.00±666.67 & 300.00±350.00           \\ 
kitchen-mixed-v0    & 100.00±111.11 & 100.00±111.11 & \textbf{255.56±99.66}  & 111.11±110.43 & \textbf{122.22±133.33}         \\ 
kitchen-new-v0      & 100.00±76.19  & 542.86±227.83 & 242.86±147.23          & 528.57±210.60 & -14.29±0.00           \\ \hline
pen-v0       & 100.00±33.60  & 158.04±41.52  & \textbf{342.47±36.92}  & 179.58±20.12  & \textbf{193.07±29.37} \\ 
hammer-v0    & 100.00±113.64 & 4.58±5.16     & \textbf{253.70±127.89} & -4.49±13.35   & \textbf{15.88±4.01}  \\ 
door-v0      & 100.00±61.89  & 119.03±48.95  & \textbf{268.38±91.77}  & 104.35±42.28  & 74.71±63.09 \\ \hline
Ant-v2              & 100.00±5.31   & 109.01±6.52   & 108.33±8.65            & 107.56±10.72  & 91.88±3.53            \\ 
Hopper-v2           & 100.00±22.02  & 120.53±29.67  & 115.67±25.69  & 150.98±32.08  & 85.48±21.42          \\ 
Walker2d-v2         & 100.00±11.90  & 130.50±23.85  & 128.98±19.94           & 117.47±12.95  & 67.80±1.57            \\ \hline
maze2d-umaze-v1     & 100.00±0.89   & 98.62±1.53    & \textbf{110.70±1.38}   & 98.66±1.37    & \textbf{112.30±1.46}  \\ 
maze2d-medium-v1    & 100.00±2.94   & 90.16±4.43    & \textbf{107.67±2.06}   & 97.71±4.78    & 93.57±7.38   \\ 
maze2d-large-v1     & 100.00±9.09   & 89.24±6.16    & \textbf{144.11±4.60}   & 90.30±8.79    & 80.50±15.13           \\ 
\end{tabular}%
\caption{Normalized values of the policy for each path under a 120K step budget. Each of the online and offlinr RL validations uses 10K steps. The value normalization is made by using a randomly policy as a value of 0 and the policy by \SAC$@\text{100K}$ as the value of 100.}
\label{tab:small_budget_different_paths_in_each_env}
\end{table*}

\newpage\clearpage
\section{Performance of Fine-tuning Phase Using IQL or PEX or CalQL in All Environments}\label{asec:ft_learn_curves}

\begin{figure*}[htb]
    \centering
    \includegraphics[width=\textwidth, height=6.8cm, keepaspectratio]{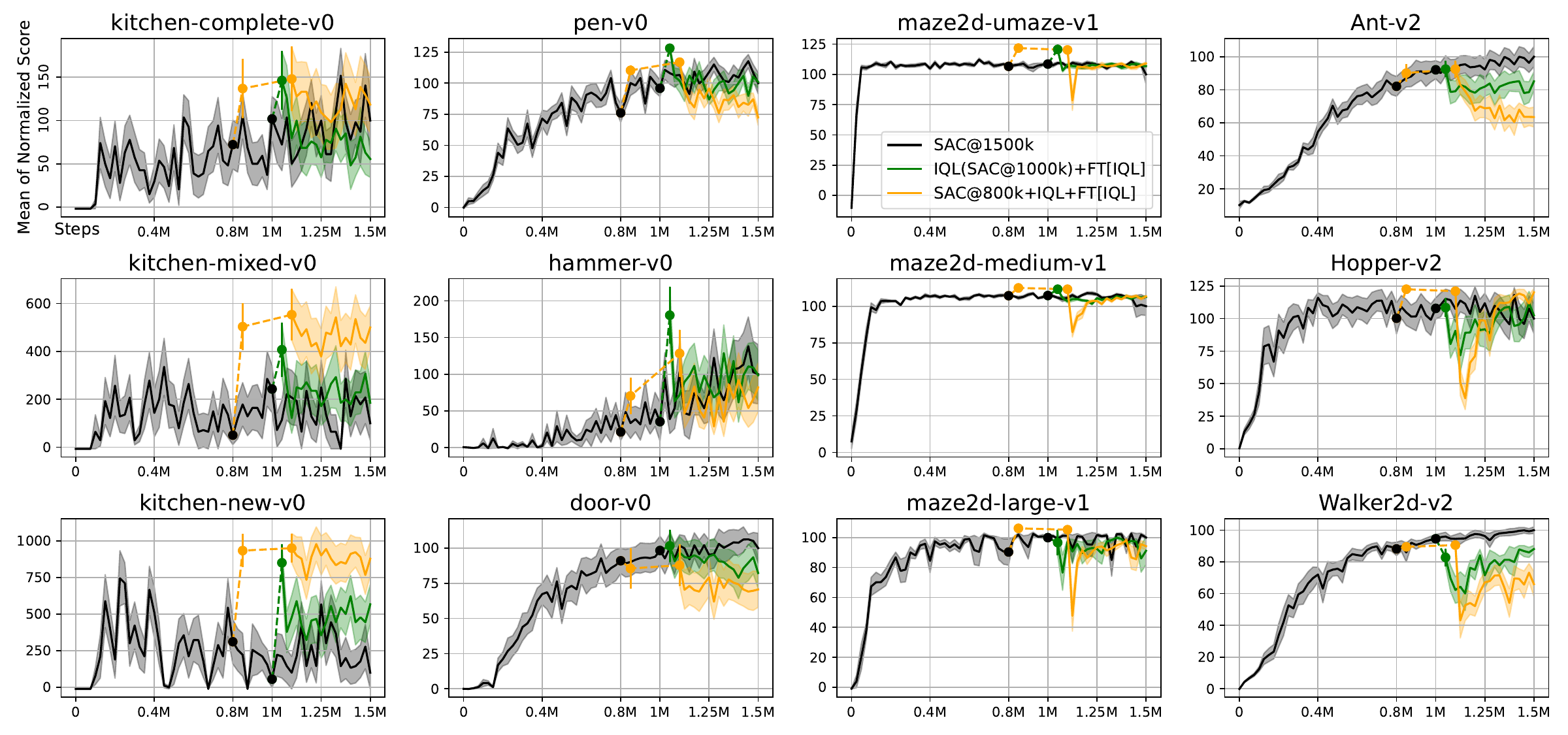}
    \caption{\SAC+\IQL+\FT{} processes with a budget of 1500K steps, where \textbf{\FT{} uses IQL}. The dashed line indicates the offline phase. For the offline phase with \textit{Buffer} strategy, \IQL(\SAC$@\text{1000K}$){}, only its output policy's score is reported; for the offline phase with \textit{NewData} strategy, \SAC$@\text{800K}$+\IQL{}, scores of both its output policy and that from \SAC$@\text{800K}${} are reported.}
    \label{fig:learning_curves_on2off2on_ftIQL}
\end{figure*}

\begin{figure*}[htb]
    \centering
    \includegraphics[width=\textwidth, height=6.8cm, keepaspectratio]{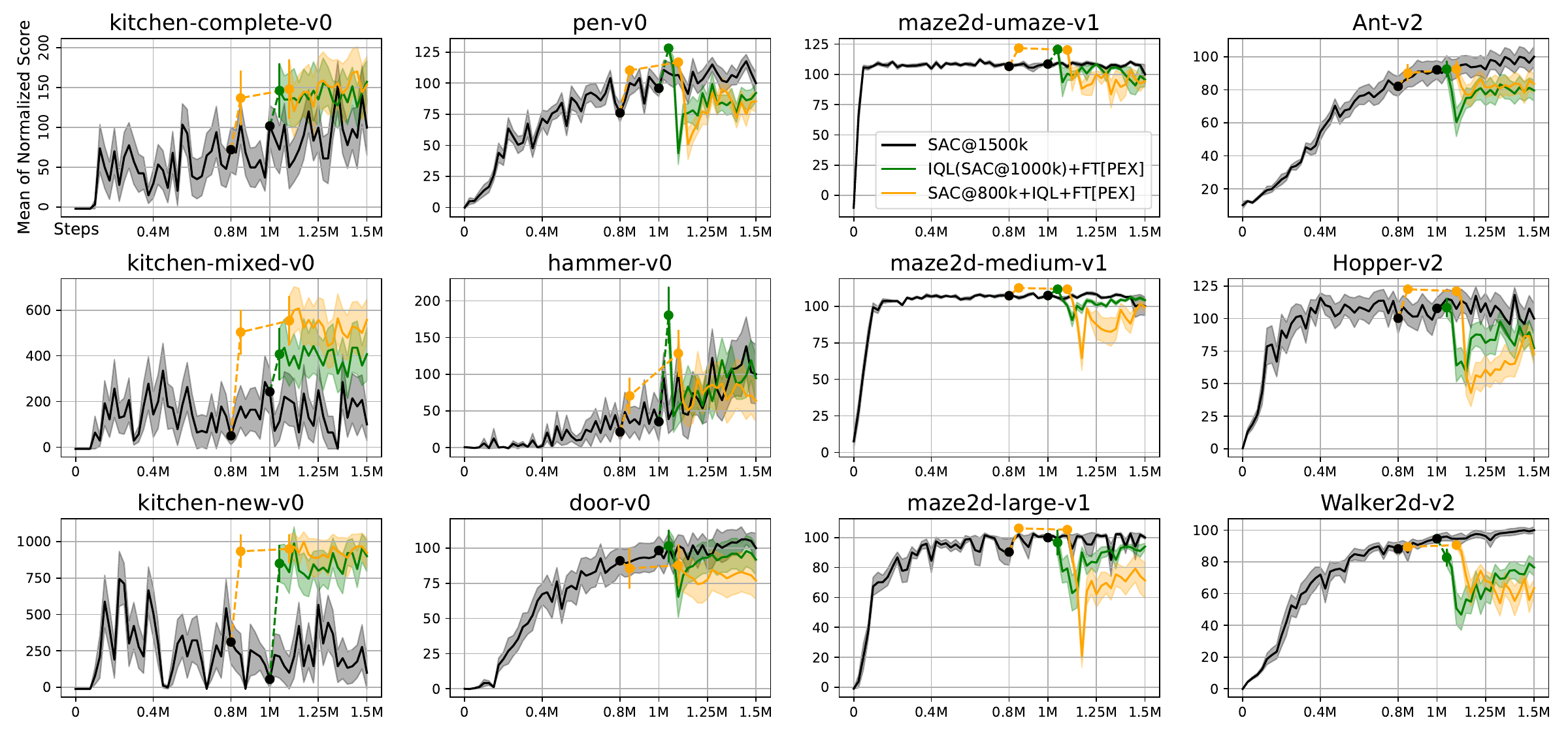}
    \caption{\SAC+\IQL+\FT{} processes with a budget of 1500K steps, where \textbf{\FT{} uses PEX}. The dashed line indicates the offline phase. For the offline phase with \textit{Buffer} strategy, \IQL(\SAC$@\text{1000K}$){}, only its output policy's score is reported; for the offline phase with \textit{NewData} strategy, \SAC$@\text{800K}$+\IQL{}, scores of both its output policy and that from \SAC$@\text{800K}${} are reported.}
    \label{fig:learning_curves_on2off2on_ftPEX}
\end{figure*}

\begin{figure*}[htb]
    \centering
    \includegraphics[width=\textwidth, height=6.8cm, keepaspectratio]{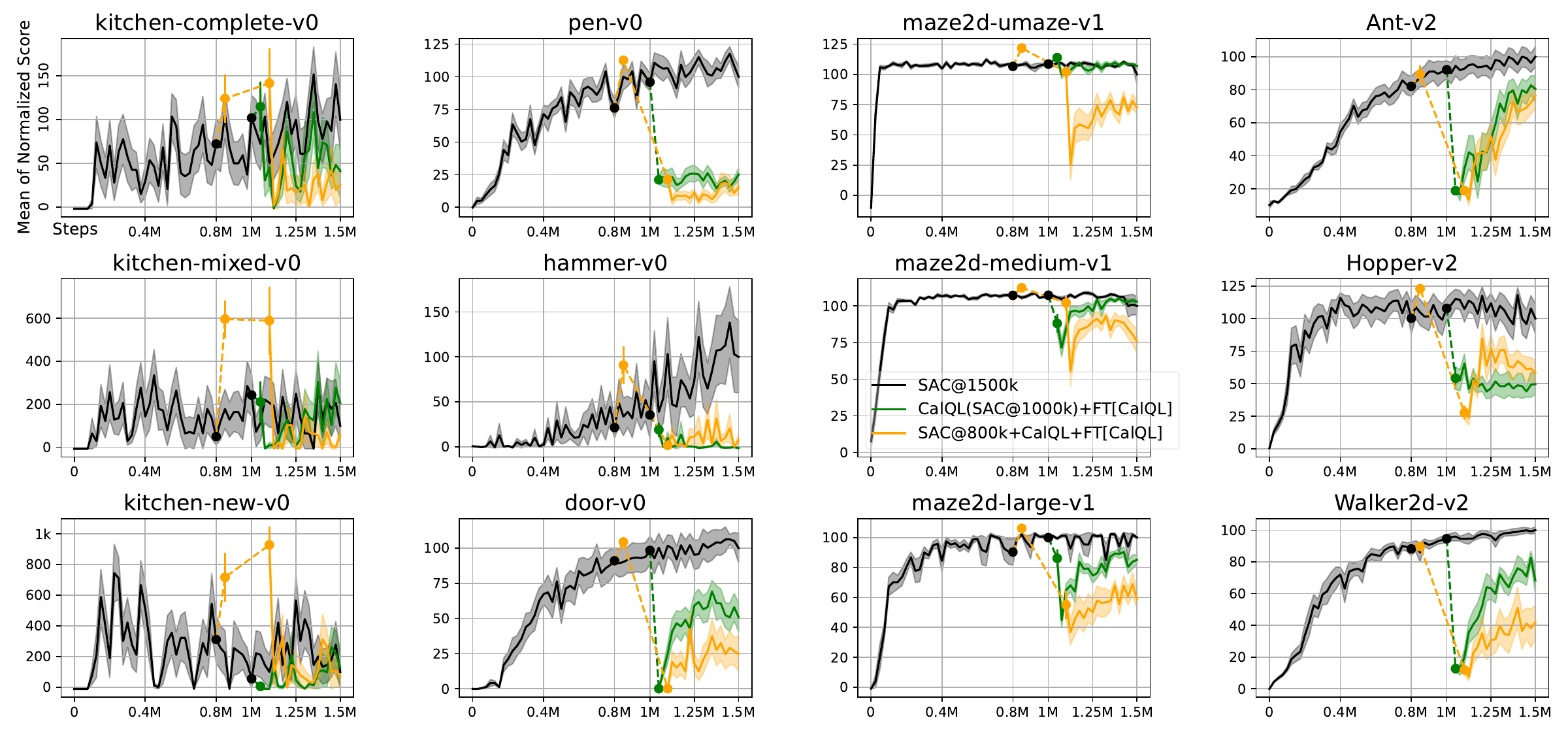}
    \caption{\SAC+\CalQL+\FT{} processes with a budget of 1500K steps, where \textbf{\FT{} uses CalQL}. The dashed line indicates the offline phase. For the offline phase with \textit{Buffer} strategy, \CalQL(\SAC$@\text{1000K}$){}, its output policy's score is reported; for the offline phase with \textit{NewData} strategy, \SAC$@\text{800K}$+\CalQL{}, scores of its output policy and that from \SAC$@\text{800K}${} are reported.}
    \label{fig:learning_curves_on2off2on_ftCalQL}
\end{figure*}

\end{document}